\documentclass{article} 
\usepackage{gmas_iclr2022_conference,times}

\usepackage{amsmath,amsfonts,bm}

\def\eqref#1{equation~\ref{#1}}

\def\1{\bm{1}}

\DeclareMathAlphabet{\mathsfit}{\encodingdefault}{\sfdefault}{m}{sl}
\SetMathAlphabet{\mathsfit}{bold}{\encodingdefault}{\sfdefault}{bx}{n}

\usepackage{algorithm}
\usepackage{algpseudocode}
\usepackage{graphicx}
\usepackage{hyperref}
\usepackage{url}

\usepackage[utf8]{inputenc} 
\usepackage[T1]{fontenc}    
\usepackage{url}            
\usepackage{booktabs}       
\usepackage{amsfonts}       
\usepackage{nicefrac}       
\usepackage{microtype}      

\usepackage{graphicx}
\usepackage[section]{placeins}
\usepackage{algorithm} 
\usepackage{algpseudocode}
\usepackage{subcaption}
\usepackage{amsmath}
\usepackage{svg}
\usepackage{comment}
\usepackage{tabularx}

\usepackage{xcolor}         
\usepackage{hyperref}       

\title{Collaborative Auto-Curricula Multi-Agent Reinforcement Learning with Graph Neural Network Communication Layer for Open-ended Wildfire-Management Resource Distribution}

\author{Philipp D. Siedler \\
Independent Researcher\\
London, UK\\
\texttt{\{p.d.siedler\}@gmail.com}
}

\iclrfinalcopy 
\begin{document}

\maketitle

\begin{abstract}
Most real-world domains can be formulated as multi-agent (MA) systems. Intentionality sharing agents can solve more complex tasks by collaborating, possibly in less time. True cooperative actions are beneficial for egoistic and collective reasons. However, teaching individual agents to sacrifice egoistic benefits for a better collective performance seems challenging. We build on a recently proposed Multi-Agent Reinforcement Learning (MARL) mechanism with a Graph Neural Network (GNN) communication layer. Rarely chosen communication actions were marginally beneficial. Here we propose a MARL system in which agents can help collaborators perform better while risking low individual performance. We conduct our study in the context of resource distribution for wildfire management. Communicating environmental features and partially observable fire occurrence help the agent collective to pre-emptively distribute resources. Furthermore, we introduce a procedural training environment accommodating auto-curricula and open-endedness towards better generalizability. Our MA communication proposal outperforms a Greedy Heuristic Baseline and a Single-Agent (SA) setup. We further demonstrate how auto-curricula and openendedness improves generalizability of our MA proposal.

\end{abstract}

\begin{figure}[!h]
\begin{center}
\vspace{-0.5cm}
\includegraphics[width=\textwidth]{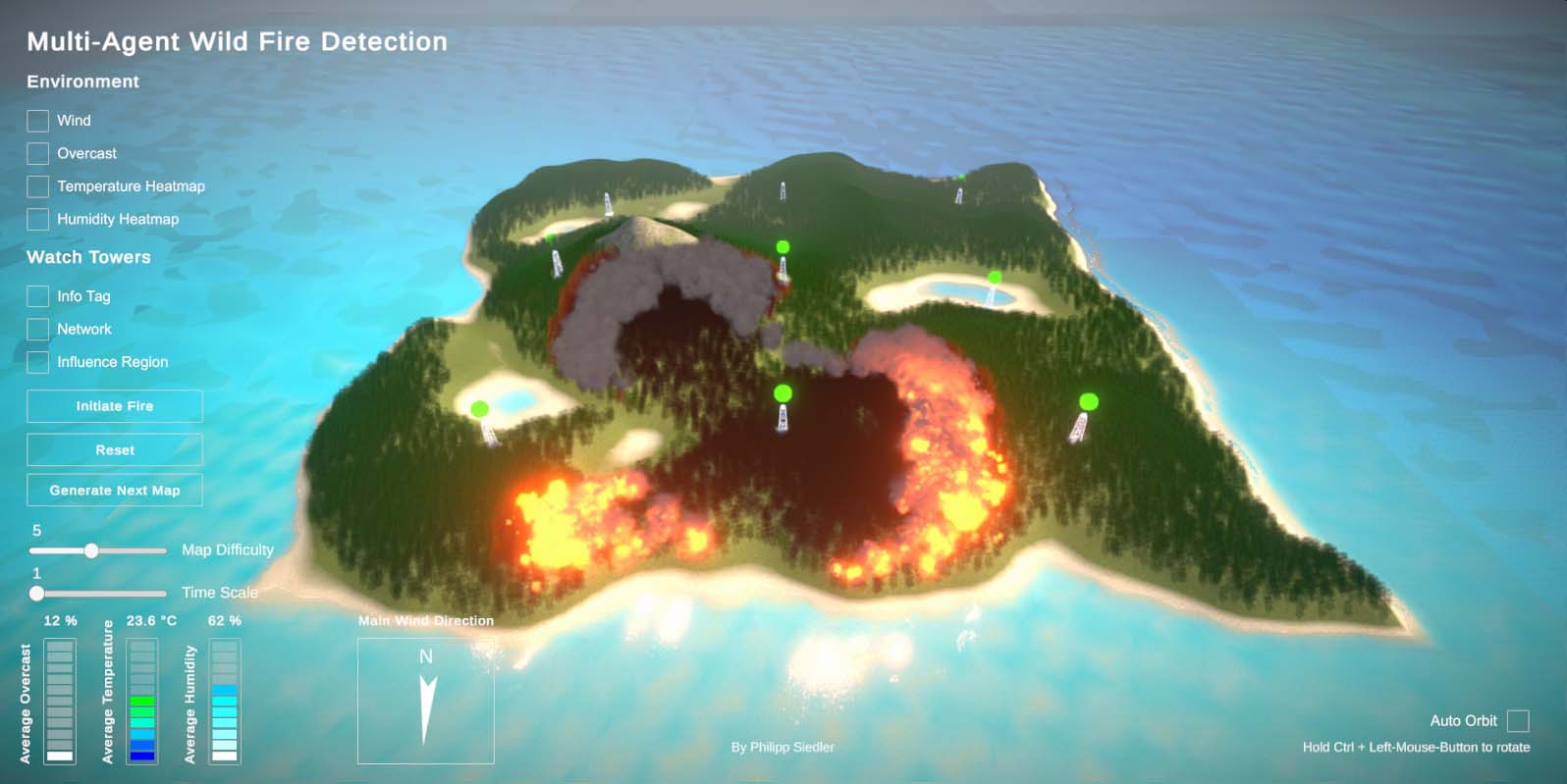}
\end{center}
\caption{Dashboard of multi-agent wildfire environment in inference mode. A web application can be found at: \url{https://philippds-pages.github.io/RL-Wild-Fire_WebApp/}.}
\end{figure}

\section{Introduction}


\subsection{Motivation}

The human ability to communicate motives and intentions is the basis for our society's  success. If the intersection of such is large enough, communication can lead to collaboration between involved parties. A collaboration exists if all collaborators behave beneficially for themselves and the collective of parties. A collective party can represent a group or an individual with an agency; therefore, we can consider entities with specific interests agents.
Many, if not all, domains in society involve multiple agents and can be described as MA systems.
While human intelligence is the highest observed in nature, not only human societies strive through collaboration.
Insects and Mammals collaborate to procreate, care, protect and ultimately survive as species. Bees, i.e. assign each other tasks, request help and fight off predators \citep{bonabeau_swarm_1999}. The highly cooperative Meerkats organise themselves in groups of up to fifty members, while some breed, others help raise the young offspring by foraging and keeping watch \citep{clutton-brock_breeding_2002}. Some monkeys and lions even kill the young of revivals, i.e. to maintain food source saturation \citep{hoogland_infanticide_1985}.

In the context of nature, Charles Darwin argues for the survival of the fittest \citep{darwin_origin_1977} and, therefore, the occurrence of competition. While in Artificial Intelligence, the majority of significant work on MA systems consider two opposing agents only, the problems of interest of this work are cooperative MA systems, where groups of agents act together to achieve higher individual and collective goals \citep{cohen_team_1997, guestrin_coordinated_2002, decker_distributed_1987, panait_cooperative_2005, mataric_using_1998}. Just like in human society or the animal world, individuals have unique or mixtures of motives. However, we can define agents with mixed or identical motives in a MA environment simulation. Assuming shared intentionality leaves us with the question of how to collaborate. Communication can play a crucial role to collaborate successfully. Human society uses language as communication medium \citep{baron_birchenall_animal_2016}. Agents can send signals of various types as a form of language. Nevertheless, observing others' behaviour can be a form of communication. Body language, a tail-wagging dog, or the red colour of an octopus can communicate internal states and intention. But we can also design agents that directly share policies - state action transitions - or memory data of past  experiences. Core questions we ask: Can agents in a MA system learn the importance of communication? Subsequently: Can agents learn to use the communicated information to take actions beneficial for themselves and the collective? And finally: Can agents learn to ask for help, form temporary alliances to encounter a high-stress state? Answering these questions will be the challenge of the experiments conducted and presented in this paper.

\subsection{Contribution}

In this work, we study communication in the context of a distributed wildfire lookout tower grid. Wildfires occur continuously and globally as part of the Earth's ecosystem \citep{bond_fire_2005}. Furthermore, climate change increases the likelihood of extreme wildfire conditions worldwide \citep{goss_climate_2020, coogan_scientists_2019}. Approximately 420 Mha is the total estimated area burned annually. While humans initiate 90\% of wildfires and only 10\% by lightning, environmental conditions, topography and fuel composition can suppress or enhance occurrence and growth. Satellites can detect fires, but only once they are already too large, and some areas are not well covered by cell phone networks. We propose unmanned lookout towers equipped with environmental sensors, cameras and local processing units. LoRa (long-range) is a low-power wide-area network modulation technique that can send signals with low power requirements in ranges of up to 15 kilometres \citep{semtech_corporation_lora_2020}. As outlined in the introduction, we chose the wildfire problem context based on its high impact and to minimise the gap between the testbed environment and the real world. By working closely on a real-world problem and its complexity, we believe we can achieve better MA collaboration systems, which is the main focus of this paper \citep{kuttler_nethack_2020}. We use a graph to organize our proposed MA lookout tower grid. Each lookout tower agent is represented by a node and has three closest neighbours to exchange local information. Such information exchange can help predict fire growth and fire management resource distribution across the lookout tower grid. Proposed communication mechanisms consist of two strands: 1. streaming of information between neighbours and 2. requesting help and answering a help request as part of the agent action space. The environment includes multiple environmental conditions, topology, and distributed fuel in the form of forest volume. To minimise simulation inaccuracies and increase generalizability, we can procedurally generate an infinite amount of environment conditions with varying difficulty levels \citep{jaderberg_open-ended_2021}. Furthermore, the environment design allows for auto-curricula, in which the MA collective can advance automatically from one difficulty level to another to improve training \citep{baker_emergent_2020}.


We present a message passing \citep{gilmer_neural_2017} GNN \citep{scarselli_graph_2009} based communication layer on top of a MARL mechanism \citep{zhang_multi-agent_2021}. We use a Proximal Policy Optimization (PPO) algorithm \citep{schulman_proximal_2017} for our Reinforcement Learning (RL) agents. We demonstrate how our collaboration mechanism using communication can help agents organise themselves to surpass a Greedy Heuristic, SA baselines. Furthermore, how our environment design, accomodating for openendedness and auto-curricula, can help our MA system advance further to become more robust and generally perform well, even in unseen environments.

\section{Related Work}
While wildfire science is not the main domain of this paper, we still think it is important to set the stage and point to some relevant work in the field we have drawn inspiration and insight from. Generally, we found that RL approaches in wildfire applications are vastly underrepresented \citep{jain_review_2020}. Only a few works use RL algorithms, such as advanced actor-critic (A3C) and Monte Carlo tree search (MCTS) methods, addressing topics related to fire behaviour prediction, more specifically fire spread and growth \citep{subramanian_learning_2017,ganapathi_subramanian_using_2018}. Nevertheless, we have been pinpointing aspects from various applications in wildfire science that help us develop a better wildfire simulation environment. One interesting aspect resonating with our communication approach is remote data sensing in possibly hard to reach terrain, spread across a network of agents organised in proximity neighbourhoods and directed graphs \citep{huot_next_2021}.
While we are looking at a highly distributed stationary MA system, there has been work on networks \citep{haksar_distributed_2018}
of autonomous unmanned aerial vehicles (UAV) to monitor and predict wildfire growth \citep{julian_image-based_2019, afghah_wildfire_2019}. 
Finally, we want to mention work on predicting wildfire using climate data, which is part of our agents sensing abilities \citep{xiong_machine_2020}.

RL is a powerful learning paradigm consisting of an agent interacting with an environment to learn from experiences through positive or negative rewards. From the perspective of an individual agent in a MA system, all other agents are part of the environment. Therefore SARL (Single-Agent Reinforcement Learning) is building the foundation for MARL \citep{wang_origin_2017}. RL does not require any data; consequently the learning environment plays a crucial role in providing enough and diverse experiences \citep{jaderberg_open-ended_2021} in conjunction with carefully crafted reward signals. Many domains are interested in RL research and applications \citep{leitao_industrial_2015}, such as game theory and distributed systems, but also optimal control, autonomous cars \citep{shalev-shwartz_safe_2016} and robotics \citep{kober_reinforcement_2013, sukthankar_cooperative_2017, ismail_survey_2018}. Games have been part of one of three main historical threads of RL development \citep{sutton_reinforcement_2015}. Therefore naturally, MARL has been studied using a diversity of games, including multiple competing and cooperating players. Traditional two-player tabletop games such as GO \citep{silver_mastering_2016, silver_mastering_2017}, Chess \citep{campbell_deep_2002}, Shogi \citep{silver_general_2018} and Hex \citep{anthony_thinking_2017}, recent work on multi-player games such as Poker \citep{moravcik_deepstack_2017, brown_superhuman_2018} and Diplomacy \citep{anthony_learning_2022, calhamer_diplomacy_1959}, but also computer games including Atari games \citep{mnih_human-level_2015}, Dota \citep{berner_dota_2019}, Starcraft \citep{vinyals_grandmaster_2019} and overcooked \citep{fontaine_importance_2021} have significantly shaped developments in AGI and RL research. Game engines such as unity include realistic physics \citep{ward_using_2020}, ideal for digital twins of real-world scenarios.

While there is work on various methods on collaboration without active communication \citep{matignon_independent_2012, panait_cooperative_2005} such as gradient-based distributed policy search \citep{peshkin_learning_2000}, reward function sharing \citep{lauer_algorithm_2000}, memory sharing \citep{lowe_multi-agent_2017, pesce_improving_2020, hernandez-leal_survey_2019} and parameter sharing (PS) \citep{sukthankar_cooperative_2017, hernandez-leal_survey_2019}, we are interested in communication as part of the agents' action space \citep{xuan_communication_2001}. Active communication requires a protocol and a medium. A protocol describes the rule of communication. The medium could be anything from low-level binary data \citep{berna-koes_communication_2004}, discrete or continuous, text, numbers-based or a combination of such as message packages. In our work, vectors of observations are part of the messages sent \citep{mataric_using_1998}, but other work proposes transferring more complex information, such as intentions or policy gradients \citep{foerster_learning_2016}. Our work can be classified as a decentralised, partially observable Markov decision process (Dec-POMDP) \citep{oliehoek_decentralized_2012} in combination with a GNN \citep{scarselli_graph_2009} message passing \citep{gilmer_neural_2017} communication layer. The proposed MA communication mechanism builds on a combination of previous work, including communication as part of the agents' action space \citep{foerster_learning_2016}, enabling the agent to send help requests to members of its neighbourhood and information broadcasting as an extension of the lookout towers local sensing capabilities \citep{sukhbaatar_learning_2016}. Our message passing GNN is structuring incoming neighbourhood information, while work by \citep{almasan_deep_2020} implemented a GNN, as part of the main training feedback. Recently published work on MARL and GNN communication layer is demonstrating how communication can improve the collective performance, and the importance of shared information \citep{siedler_power_2021}. Here we advance the communication protocol and the agents ability to raise and answer help requests actively. Requesting help, as well as helping is part of the agents action space.

\section{Background}

\subsection{Proximal Policy Optimisation}

All agents are trained using state of the art algorithm Proximity Policy Optimization (PPO). Two main concepts distinguish PPO. Firstly, PPO estimates a trust region to take safe learning steps while performing gradient ascent. Secondly, Advantage estimates how good an action is compared to the average action in a specific state. Many other RL algorithms, such as Asynchronous Advantage Actor Critic (A3C), use this concept \citep{udacity-deeprl_introduction_2019}.
\textbf{Advantage:} Advantage can be described as the difference of the Q Function and the Value Function: $A(s,a) = Q(s,a) - V(s)$, where $s$ is the state and $a$ the action \citep{zychlinski_complete_2019}.
The Q Value (Q Function), denoted as $Q(s,a)$,
measures the overall expected reward given state $s$, performing action $a$. Assuming the agent continues playing until the end of the episode following policy $\pi$. The Q is abbreviated from the word Quality, and denoted as: $\mathcal{Q}(s,a) = \mathbb{E}\left[ \sum_{n=0}^{N} \gamma^n r_n \right]$.
The State Value Function, denoted as $V(s)$, measures, similar to the Q Function, overall expected reward, with the difference that the State Value is calculated after the action has been taken and is denoted as: $\mathcal{V}(s) = \mathbb{E}\left[ \sum_{n=0}^{N} \gamma^n r_n \right]$. The Q Value $V(s)$, with $n=0$, is the expected reward $r^0$ in state $s$, before action $a$ was taken, while the Q Value measures the expected reward $r^0$ after $a$ was taken.
\textbf{Trust Region:} After some experience samples $\pi_{\theta_k}(a_t|s_t)$ have been collected, the trust region can be calculated as the quotient of the current policy to be refined $\pi_\theta(a_t|s_t)$ and the previous policy as follows $r_t(\theta) = \frac{\pi_\theta(a_t|s_t)}{\pi_{\theta_k}(a_t|s_t)} = \frac{current\ policy}{old \  policy}$.
This is a simplified gradient ascent objective function with limited deviation between the current and old policies \citep{achiam_simplified_2018}.
\begin{quote}
$\mathcal{L}_{\theta_k}^{CLIP}(\theta) = \underset{s,a\sim\theta_k}{\mathbb{E}} \left[\min{\left( r_t(\theta)A^{\theta_k}(s,a), g(\epsilon,A^{\theta_k}(s,a))\right)}\right]$,
\newline
\newline
where
\newline
\newline
$g(\epsilon,A) = 
\begin{cases}
(1 + \epsilon)A,& \text{if } A\geq 0\\
(1 - \epsilon)A,& \text{otherwise}
\end{cases}
$
\end{quote}
The advantage function will be clipped to the value at ($1-\epsilon$) or ($1+\epsilon$), if the probability ratio between the current and the previous policy is outside the range of ($1+\epsilon$) and ($1-\epsilon$). This also means that the advantage will never exceed the clipped values. In the original PPO paper by \citep{schulman_proximal_2017} $\epsilon$ was set to 0.2.
Finally, the policy that yields the highest sum over all Advantage estimates $A_t$ in range of max time step $T$ of a trajectory $\tau \in \mathbb{D}_k$ will be used to override the old policy $\theta_{old}$ \citep{openai_proximal_2021}: $\theta_{k+1} = arg \underset{\theta_k}{max}\frac{1}{|\mathbb{D}_k|T}\sum_{\tau \in \mathbb{D}_k}\sum_{t=0}^{T}\min\left( \frac{\pi_\theta(a_t|s_t)}{\pi_{\theta_k}(a_t|s_t)},g(\epsilon,A^{\theta_k}(s,a)) \right)$.

\subsection{Graph Neural Network}
\begin{figure}[!ht]
\begin{subfigure}{0.245\textwidth}
\centering
\includegraphics[width=0.8\linewidth]{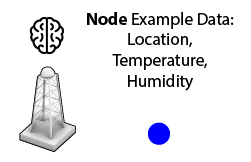}
\caption{Graph Node}
\label{fig:graph-node}
\end{subfigure}
\begin{subfigure}{0.245\textwidth}
\centering
\includegraphics[width=0.8\linewidth]{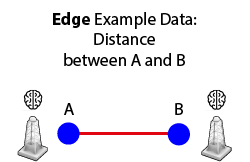}
\caption{Graph Edge}
\label{fig:graph-edge}
\end{subfigure}
\begin{subfigure}{0.245\textwidth}
\centering
\includegraphics[width=0.8\linewidth]{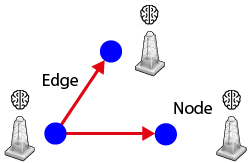}
\caption{Directed Graph}
\label{fig:directed-graph}
\end{subfigure}
\begin{subfigure}{0.245\textwidth}
\centering
\includegraphics[width=0.8\linewidth]{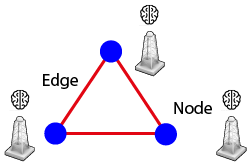}
\caption{Undirected Graph}
\label{fig:undirected-graph}
\end{subfigure}
\caption{Graph $\mathcal{G}$ consisting of vertices $\mathcal{V}$ (blue dots) and edges $\mathcal{E}$ (red lines): $\mathcal{G} = (\mathcal{V},\mathcal{E})$}
\end{figure}

Many flavours of GNNs exist \citep{li_gated_2017, velickovic_graph_2018, defferrard_convolutional_2017}, but \citep{scarselli_graph_2009} fundamentally introduced them.
A graph is a data structure based on nodes or vertices and edges. Nodes are objects holding arbitrary features. Edges represent the relationships between nodes. Edges can be directed from node A to B \ref{fig:directed-graph}, or undirected, from node A to B and vice versa \ref{fig:undirected-graph}.

The basic functionalities of GNNs are graph, node and edge classification. Features of a node can be predicted using edges or the existence of edge connections using node features. Graphs as a whole can be classified i.e. using node features and the graphs topology. However, the simplest form of a GNN is the message passing framework proposed by \citep{gilmer_neural_2017} using the network architecture introduced by \citep{battaglia_relational_2018}, utilising a "graph-in, graph-out" architecture. The input graph topology is not modified but its loaded feature embeddings.

Node states can be denoted as $v$, edges connecting with node $v$ as $x_{co[v]}$. The state of a node $h_v$ consists of n-dimensional vector features. Adjacencies between a node and its neighbours are the mapped transition of the node, denoted as $h_{ne[v]}$, including all neighbouring node features, denoted as $x_{ne[v]}$.
The transition function $f$ is used to embed each node on a n-dimensional space \citep{zhou_graph_2020}: $h_v = f(x_v, x_{co[v]}, h_{ne[v], x_{ne[v]}})$ \\
While the two most popular algorithms to define neighbourhoods on graphs are Breadth-First Search (BFS) \citep{burkhardt_optimal_2021}, Depth-First Search (DFS) \citep{kaur_analysis_2012} and random walk based DeepWalk \citep{perozzi_deepwalk_2014}, we define neighbourhoods by finding Euclidean distance based $n$ nearest neighbours.
Passing state $h_v$ and feature $x_v$ to the GNN outputs the result of function g: $o_v = g(h_v, x_v)$.
A basic last step is applying gradient descent to formulate loss using the ground truth $t_v$ as well as the output $o_v$ of node $v$: $loss = \sum_{i=1}^{p}(t_i - o_i)$. In our approach we are using gradient ascent and a reward function utilised by PPO.
  
\subsection{Multi-Agent Communication}

\begin{figure}[!h]
\includegraphics[width=\linewidth]{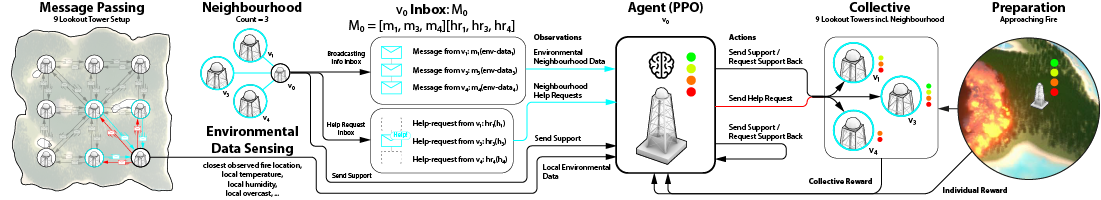} 
\caption{GNN Message Passing Communication Diagram: Neighbourhood graph (n=3); Observations: Inbox environmental data, inbox help requests and local environmental data; Agent (PPO); Actions: send support, request support back, send a help request to others, support self; Rewards: individual and collective reward for preparedness in case of fire. A zoomed-in version of this diagram can be found in the appendix: Figure \ref{fig:communication_diagram_large}.}
\label{fig:GNN_MessagePassing}
\end{figure}

Agent communication is possible between an agent and the agents in its neighbourhood. Two main functionalities define communication. Firstly, each agent can send help request $hr$ messages as part of its action space. Help request messages are sent at time step $t$ and received by all neighbours $u_v$ of $v$ at time step $t+1$. The neighbouring agents can now collaborate and react to the help request $hr_t$ by sending resources, if available, at time step $t+2$ to support $v$. If the sent resources help $v$, $u_v$ gets a small positive bonus reward of $+0.1$ for helping, no other agent can receive a bonus reward for reacting to ${hr}_t$ thereafter. A help request message consists of a boolean signal, $hr_t = [false/true]$. An agent can receive multiple help requests as part of its help request inbox $hri[{hr}_1, {hr}_2, {hr}_3]$ and needs to decide which to react to if at all. And secondly, information broadcasting between each agent $v$ and its neighbours $u_v$. All neighbours receive broadcasted messages in an inbox $ibi[{ib_1}, {ib_2}, {ib_3},]$. Received messages $ib[{cof}_{pos}(x, y, z), temp, hum, prep, oc]$ including information such as closest observed fire location (if existing) ${cof}_{pos}$, in the form of a 3-dimensional vector ${cof}_{pos}(x, y, z)$, local temperature $temp$, humidity $hum$, the current preparation value $prep$ and the percentage of overcast $oc$ at the lookout tower location as scalars. The graph-structured communication is the input to the neural network of the GNN. The agent has to learn how to reason about the broadcasting information and whether to support its neighbours or itself in preparation for approaching fire. While helping a neighbour might yield a bonus reward, there is a risk for unpreparedness of its own lookout tower, which might result in low rewards.

\section{Methodology}

\subsection{Environment}
\begin{figure}[!ht]
\begin{subfigure}{0.19\textwidth}
\centering
\includegraphics[width=\textwidth]{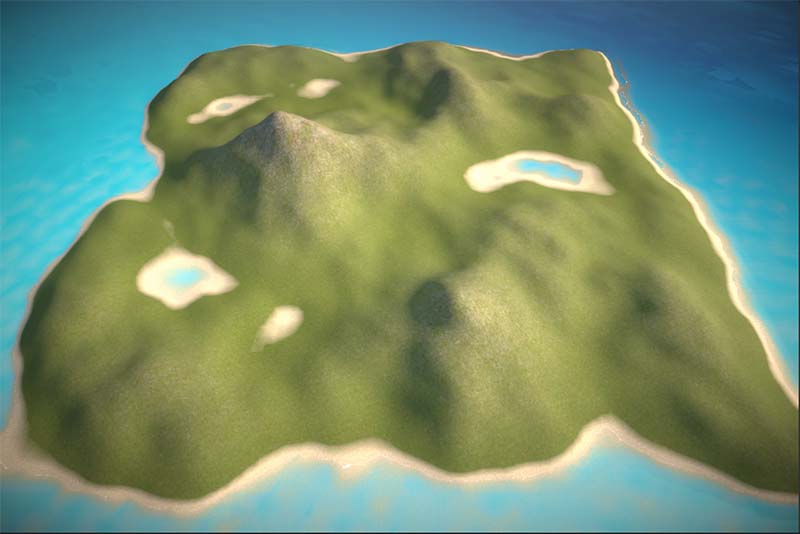}
\caption{Terrain}
\label{fig:static-env-terrain}
\end{subfigure}
\begin{subfigure}{0.19\textwidth}
\centering
\includegraphics[width=\textwidth]{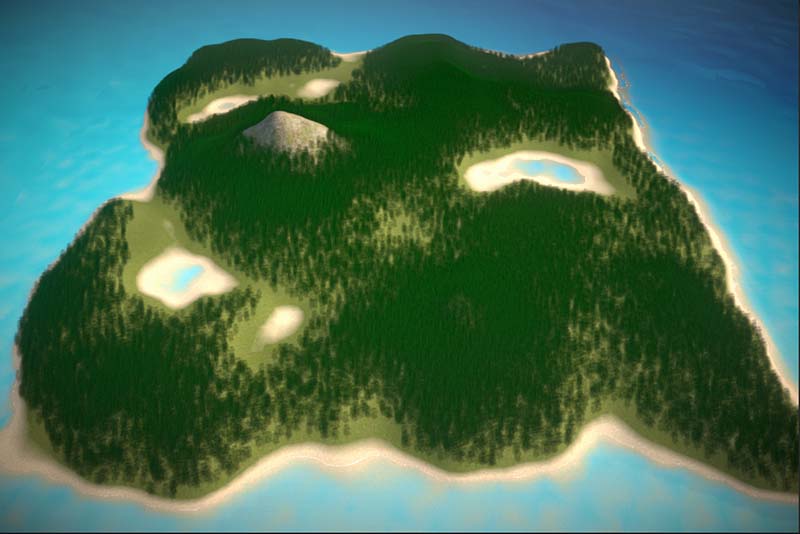}
\caption{Forest}
\label{fig:static-env-forest}
\end{subfigure}
\begin{subfigure}{0.19\textwidth}
\centering
\includegraphics[width=\textwidth]{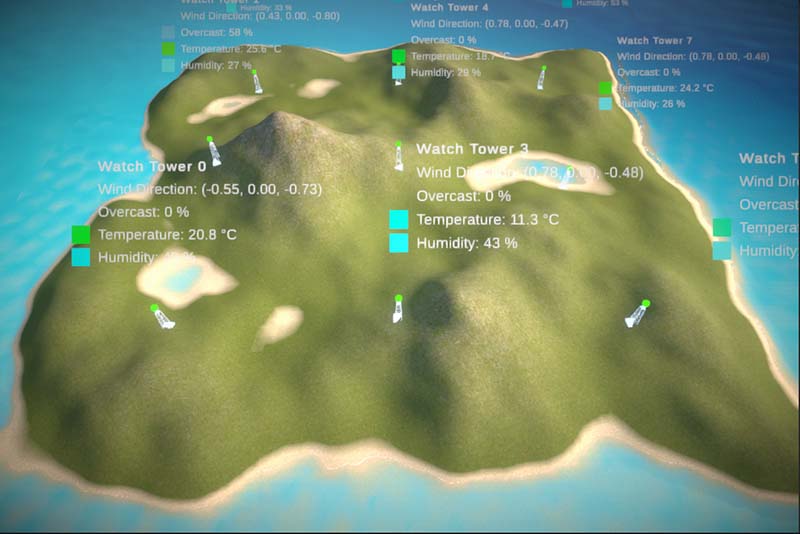}
\caption{Lookout Towers}
\label{fig:static-env-wt}
\end{subfigure}
\begin{subfigure}{0.19\textwidth}
\centering
\includegraphics[width=\textwidth]{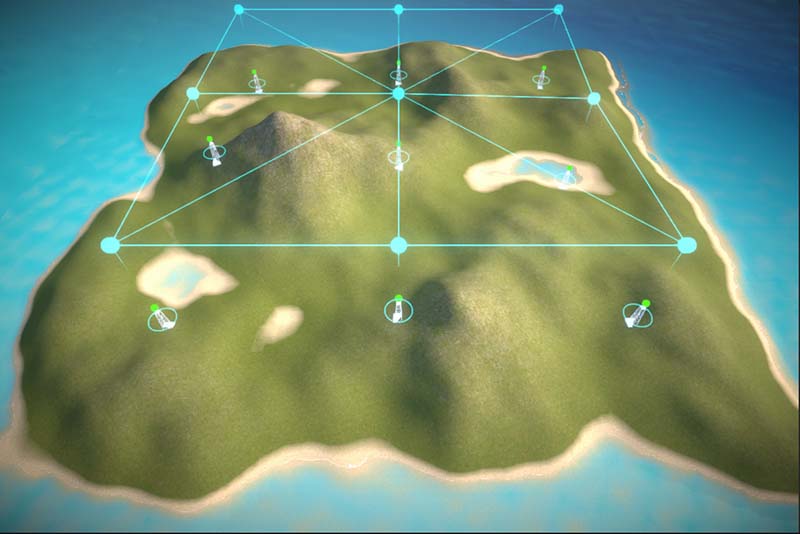}
\caption{Network}
\label{fig:static-env-network}
\end{subfigure}
\begin{subfigure}{0.19\textwidth}
\centering
\includegraphics[width=\textwidth]{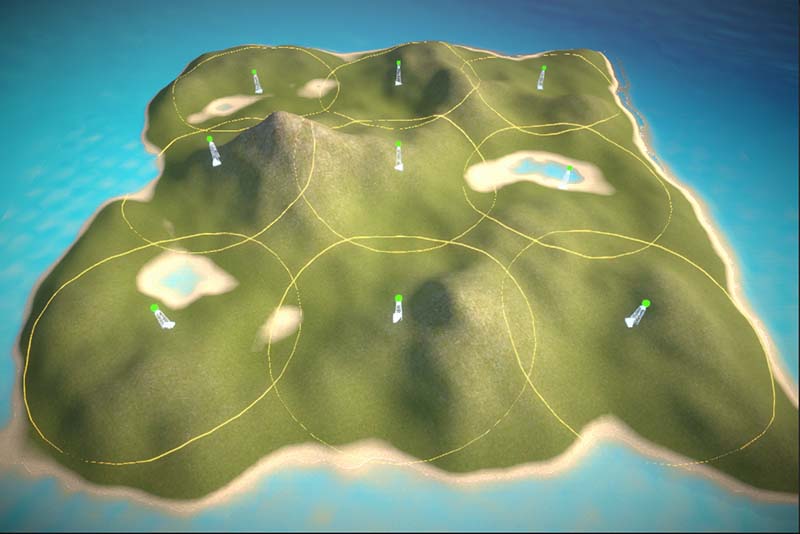}
\caption{Lookout Region}
\label{fig:static-env-influence}
\end{subfigure}
\caption{Static environment features. Zoomed-in version in the appendix: \ref{appendix:static-and-dynamic-env-1}.}
\label{fig:static-env}
\end{figure}

We now describe the 3-D Wildfire Lookout Tower environment, developed in the game engine unity, used for training and evaluating the Greedy Heuristic Baseline, Single- and Multi-Agent experiments. The scenario is a procedurally generated landmass with a distributed network of lookout towers, forest and environmental condition features. Static features of the environment are shown in Figure \ref{fig:static-env}. A sample of the terrain is shown in Figure \ref{fig:static-env-terrain}. Terrain height values influence the distribution of trees forming the forest volume (\ref{fig:static-env-forest}). Nine lookout towers are distributed in a fixed three by three grid. The placement of a lookout tower also determines the neighbourhood, consisting of the three closest lookout towers. Each lookout tower has a fixed observation region (Figure \ref{fig:static-env-influence}, \ref{fig:reward_function_detail}).

\begin{figure}[!ht]
\begin{subfigure}{0.19\textwidth}
\centering
\includegraphics[width=\textwidth]{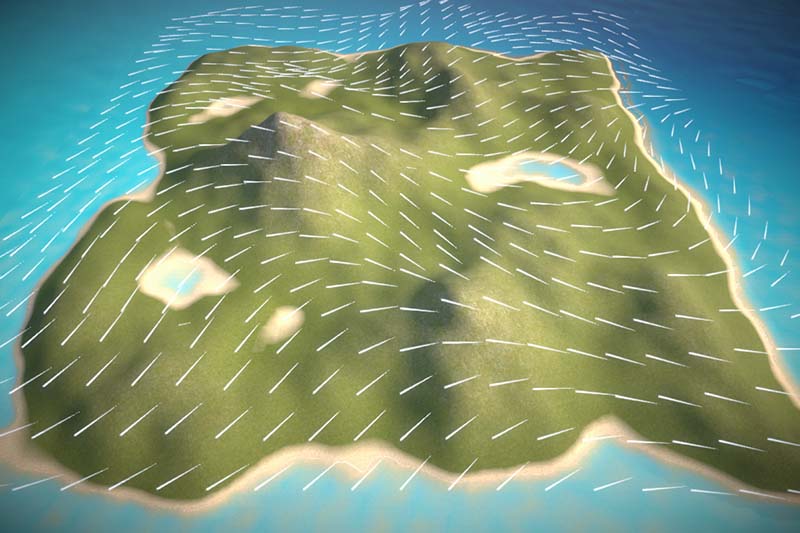}
\caption{Wind}
\label{fig:dynamic-env-wind}
\end{subfigure}
\begin{subfigure}{0.19\textwidth}
\centering
\includegraphics[width=\textwidth]{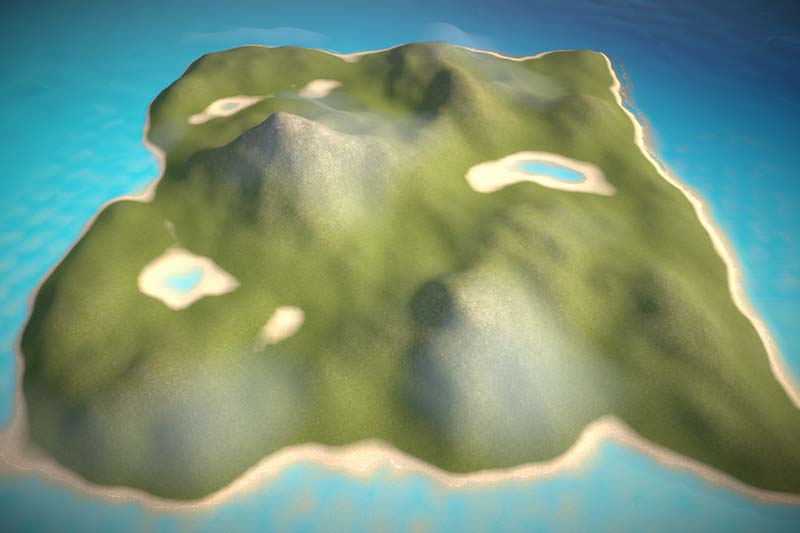}
\caption{Overcast}
\label{fig:dynamic-env-overcast}
\end{subfigure}
\begin{subfigure}{0.19\textwidth}
\centering
\includegraphics[width=\textwidth]{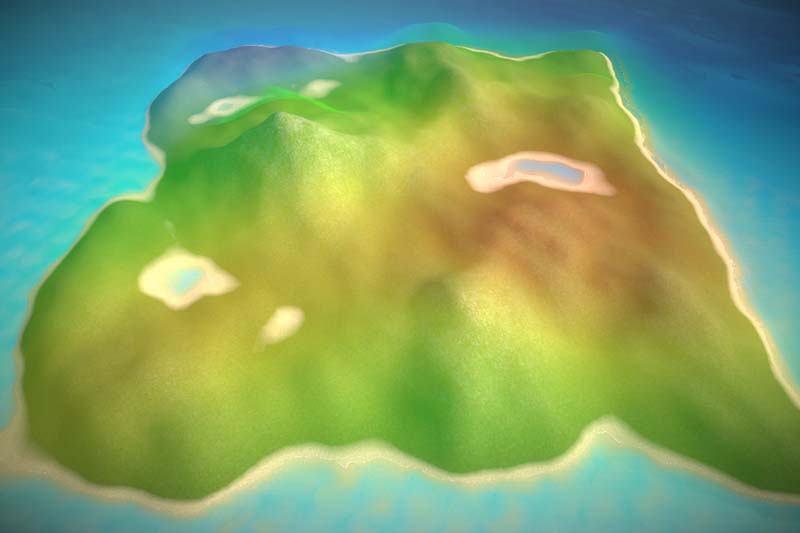}
\caption{Temperature}
\label{fig:dynamic-env-temp}
\end{subfigure}
\begin{subfigure}{0.19\textwidth}
\centering
\includegraphics[width=\textwidth]{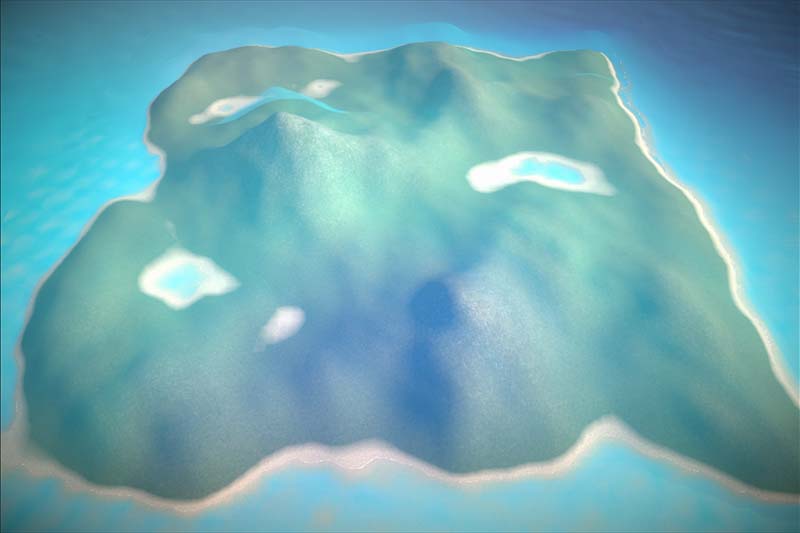}
\caption{Humidity}
\label{fig:dynamic-env-humidity}
\end{subfigure}
\begin{subfigure}{0.19\textwidth}
\centering
\includegraphics[width=\textwidth]{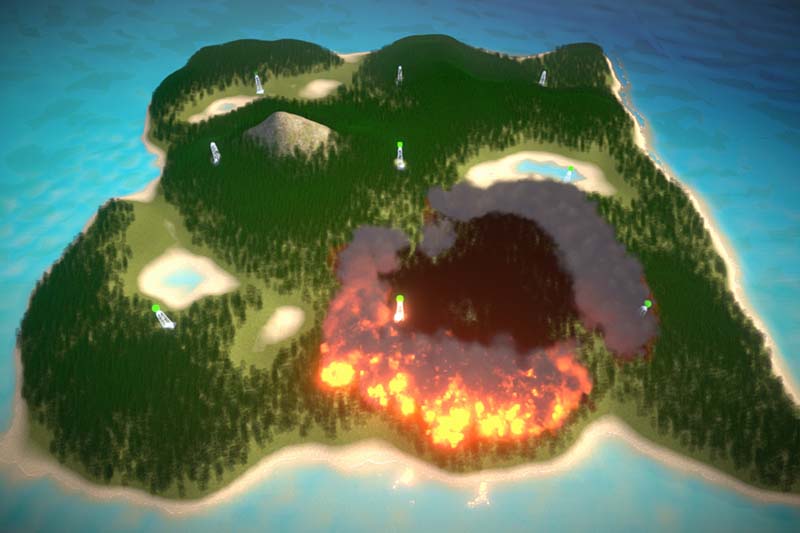}
\caption{Fire}
\label{fig:dynamic-env-fire}
\end{subfigure}
\caption{Dynamic environment features. Zoomed-in version in the appendix: \ref{appendix:static-and-dynamic-env-2}.}
\label{fig:dynamic-env}
\end{figure}

The dynamic features of the environment (Figure \ref{fig:dynamic-env}) are based on perlin noise \citep{perlin_image_1985} and a main wind direction, including a wind field (\ref{fig:dynamic-env-wind}), overcast (\ref{fig:dynamic-env-overcast}), temperature (\ref{fig:dynamic-env-temp}) and humidity (\ref{fig:dynamic-env-humidity}). Wildfire is also part of the dynamic environment features. The fire's initiation and growth are based on environmental features and probability. At the beginning of an episode the location on the terrain with the lowest overcast, highest temperature and lowest humidity is chosen to ignite a wild fire. Fire can not spread to the next tree when the distance is larger than ten meters. However, if the distance is lower, the following conditions add twenty percent each to the probability of fire spreading: if the angle between the wind direction vector and the target vector is lower than 45 degrees; if the target is at a higher location; if the target temperature is higher than 21 Celsius degree; if the target humidity is higher than 50 percent and if overcast is zero. When all conditions are true, the chance of fire spreading to the target is 100 percent. Once the fire has spread, a tree will burn ten time-steps. At this point it is important to stress that the focus of this paper is not on simulating hyper-realistic environmental conditions or fire behaviour, rather a close adjustable abstraction.

\begin{figure}[!ht]
\begin{subfigure}{0.19\textwidth}
\centering
\includegraphics[width=\textwidth]{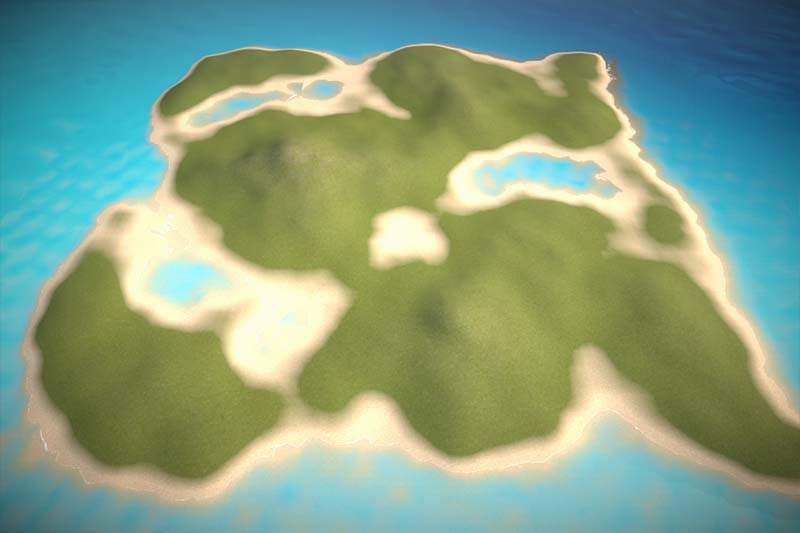}
\caption{seed=0, diffic.=1}
\label{fig:open-ended-diffic-1}
\end{subfigure}
\begin{subfigure}{0.19\textwidth}
\centering
\includegraphics[width=\textwidth]{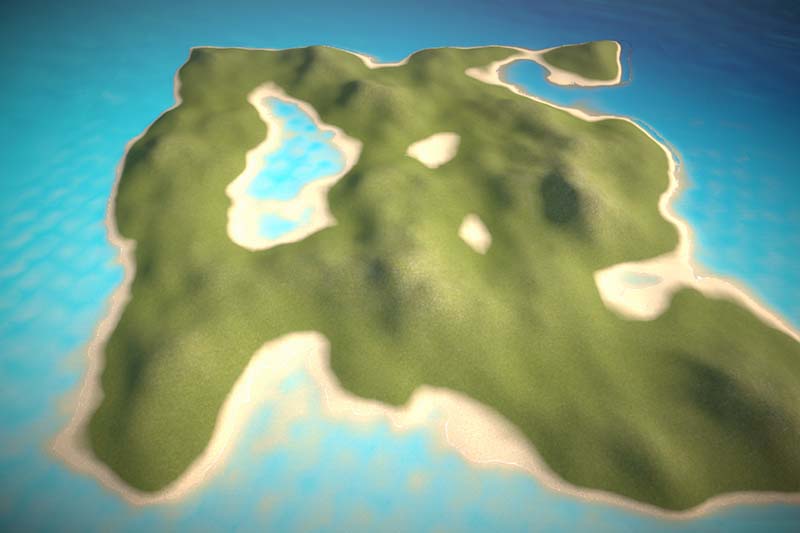}
\caption{seed=1, diffic.=3}
\label{fig:open-ended-diffic-3}
\end{subfigure}
\begin{subfigure}{0.19\textwidth}
\centering
\includegraphics[width=\textwidth]{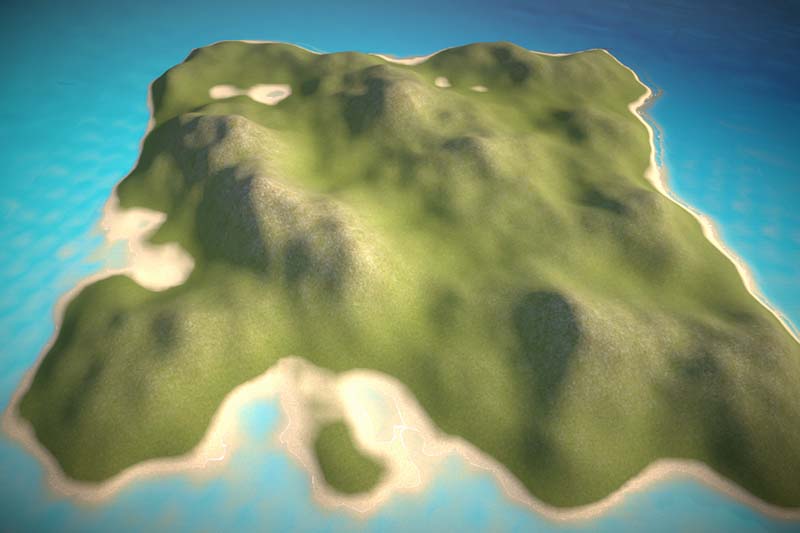}
\caption{seed=2, diffic.=5}
\label{fig:open-ended-diffic-5}
\end{subfigure}
\begin{subfigure}{0.19\textwidth}
\centering
\includegraphics[width=\textwidth]{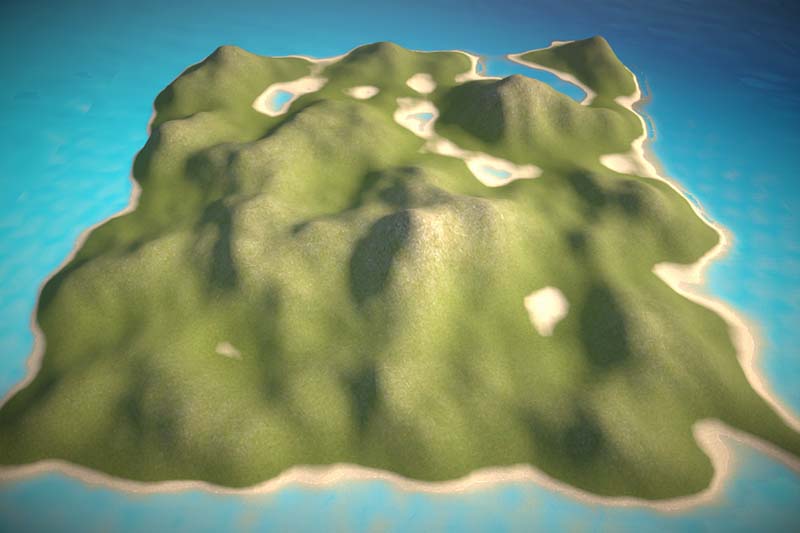}
\caption{seed=3, diffic.=7}
\label{fig:open-ended-diffic-7}
\end{subfigure}
\begin{subfigure}{0.19\textwidth}
\centering
\includegraphics[width=\textwidth]{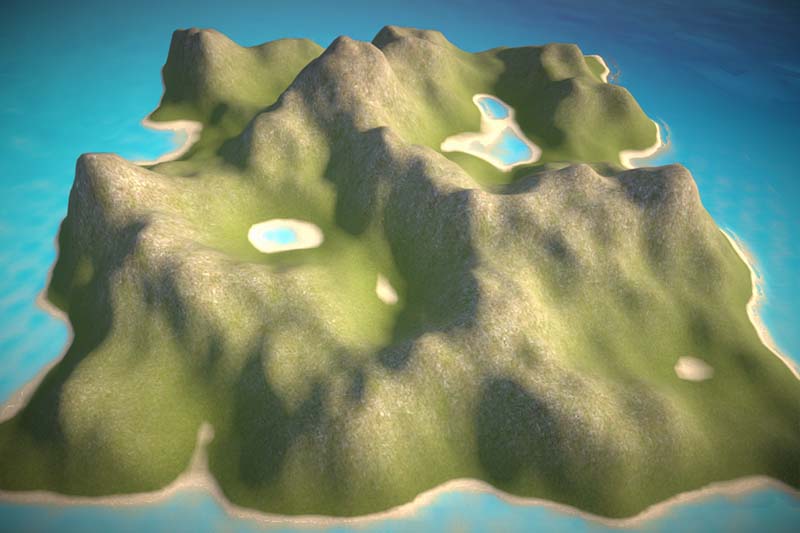}
\caption{seed=4, diffic.=10}
\label{fig:open-ended-diffic-10}
\end{subfigure}
\caption{Open-ended environment and increasing difficulty. Zoomed-in version in the appendix: \ref{appendix:difficulty-vs-seed-matrix}.}
\label{fig:open-ended-diffic}
\end{figure}

While we are aware of the fact that simulation will always differ from the real world, we try to close the simulation to reality gap by following a strategy explained by Thore Graepel, creating as much diversity in our simulation environment as possible \citep{fry_deepmind_2022}. We aim towards achieving open-endedness through procedurally generating a virtually infinite amount of terrain scenarios. Additionally, ten difficulty levels modify the height of the terrain, resulting in less observable regions for each agent and, therefore, higher communication necessity and overall higher difficulty predicting fire growth. Furhter terrain samples can be found in the difficulty vs seed matrix in the Appendix (\ref{appendix:difficulty-vs-seed-matrix}).

\subsection{Performance Evaluation Method:}

Each lookout tower has a resource reserve $rr$ of value 1.0. Initially, all lookout towers support value $sv$ is at 0.0. Resources from the resource reserve $rr$ can be distributed in 0.1 increments. Distribution targets can be self or lookout towers in the neighbourhood. In order to distribute resources to a target, the resource reserve $rr$ needs to be larger or equal to $0.1$. If the resource reserve is empty, resources have to be deducted in 0.1 decrements from self or lookout towers in the neighbourhood to free up resources for redistribution. If there is no self-need for resources, agents can collect reward fractions for distributing resources to neighbouring lookout towers in need. An example scenario could be: 0.5 resources are distributed to self, and 0.5 resources at a neighbouring lookout tower. There is no observed fire near self, but fire is approaching at the neighbouring location. The agent now gets 0.5 times the performance of the neighbouring tower, but none for the distributed resources at target self. Performance is calculated using a broken power law function using $\beta = -1$, $s = 2$, ${x_n} = 270$, $a = 5$ and $x = \frac{\text{distance to closest observed fire}}{\text{influence region distance}}$ leading to $F(x, {x_n}, a, s, \beta) = {\left(1 + {[\frac{x * 1000}{x_n}]}^{a}\right)}^{\frac{-1}{s}}$, where\\
$x =
\begin{cases}
x \text{ remapped from domain 0 to 1, to domain 0.5 to 0},& \text{if fire moves towards tower} \\
x \text{ remapped from domain 0 to 1, to domain 0.5 to 1},& \text{otherwise.}
\end{cases}$\\
The reward function will yield the highest possible reward if a tower is prepared well and has a high support value before fire crossing the influence region. Each tower relies on environmental data and observed fire locations to predict how much support preparation is needed.

\subsection{Experiments}

\begin{table}[H]
\caption{Experiment setups and parameters for lookout tower grids of 9. Auto-curriculum (AC), seed (s), environment terrain with seed 0 (s=0); infinite environment terrain scenarios (s=inf).}
\label{setup-table}
\centering
\begin{tabularx}{\textwidth}{p{3.3cm} p{0.9cm} p{0.9cm} p{1.4cm} p{1.1cm} p{0.9cm} p{1cm} p{1.4cm}}
\toprule
\multicolumn{4}{c}{} & \multicolumn{3}{c}{Observation(s)} \\
\cmidrule(r){5-7}
Setup & Agent Count & Tower Count & Neighbour Count & At each Tower & Total & Stack Size & Action(s) per Agent \\
\midrule
Greedy Heuristic            & 0 & 9 & 0 & 0 & 0 & 0 & 0 \\
Single-Agent (s=0)          & 1 & 9 & 3 & 7 & 63 & 2 & 36 \\
Multi-Agent (s=0)           & 9 & 9 & 3 & 32 & 32  & 2 & 5 \\
Multi-Agent (s=inf,AC)      & 9 & 9 & 3 & 32 & 32  & 2 & 5 \\
\bottomrule
\end{tabularx}
\end{table}

All agent experiments have been trained and tested for 500 time-steps per episode. Each agent can take multiple decisions at every time step, depending on its action space. \textbf{Greedy Heuristic Baseline:} The hand-designed greedy heuristic baseline keeps all resources to itself and does not support neighbours. Naturally, no training is required. \textbf{Single-Agent (seed 0):} The agent in the single-agent setup controls resource distribution of all lookout towers. The egoistic reward is gained by distributed resources, multiplied by performance value $p$, and the average collective reward $cr$ consisting of the average performance $ap$ over all lookout towers. The environment scenario with seed 0 is used for training. \textbf{Multi-Agent (seed 0):} In the multi-agent setup each lookout tower is controlled by an individual agent. All agents receive egoistic and collective rewards, as explained in the single-agent setup description. Additionally in the multi-agent setup, reacting to a help request first yields a small bonus reward. The environment scenario with seed 0 is used for training. \textbf{Multi-Agent (Openended \& Autocurricula):} Finally, to show the strength of openendedness (seed=inf) and auto-curricula (AC), we trained a multi-agent setup, but with changing environment scenario for each episode as well as rising difficulty level. The difficulty level advances if the agent has achieved a certain cumulative reward threshold, over 100 past episodes. Further details on the curricula design can be found in the Hyperparameters Appendix \ref{appendix:multi-agent-auto-curricula}.

\subsection{Results}

\begin{table}[H]
\caption{Experiment results while training and inference mode, including training time. Inference: Mean reward and performance for environment with seed 0 and seed inf.}
\label{performance-table}
\centering
\begin{tabularx}{\textwidth}{p{3.3cm} p{1.49cm} p{1.6cm} p{1.5cm} p{1.5cm} p{1.5cm}}
    \toprule
    \multicolumn{1}{c}{}    & \multicolumn{1}{c}{Training Time} & \multicolumn{2}{c}{Inference: Mean Reward} & \multicolumn{2}{c}{Inference: Performance} \\
    \cmidrule(r){2-2}
    \cmidrule(r){3-4}
    \cmidrule(r){5-6}
    Setup                   & 5e7 step(s)                    & seed=0                        & seed=inf                      & seed=0 & seed=inf \\
    \multicolumn{1}{c}{}    & \multicolumn{1}{l}{(↓ better)}    & \multicolumn{1}{l}{(↑ better)}    & \multicolumn{1}{l}{(↑ better)}    & \multicolumn{1}{l}{(↑ better)}    & \multicolumn{1}{l}{(↑ better)} \\
    \midrule
    Greedy Heuristic        & -                   & 122.3±46.7              & 111.2±36.1            & 0.117±0.039           & 0.109±0.034 \\
    Single-Agent (s=0)      & 787e3(sec)          & \textbf{1073.1±457.3}   & 589.9±283.3           & \textbf{0.189±0.070}  & 0.111±0.043 \\
    Multi-Agent (s=0)       & 132e3(sec)          & 995.5±172.2             & 876.1±131.2           & 0.178±0.035           & 0.154±0.226 \\
    Mutli-Agent (s=inf,AC)  & \textbf{128e3}(sec) & 967.5±164.17            & \textbf{907.0±142.7}  & 0.171±0.034           & \textbf{0.158±0.027} \\
    \bottomrule
\end{tabularx}
\end{table}

Results show that our Multi-Agent proposal surpasses the Greedy Heuristic, Single-Agent and Multi-Agent setup in unseen environments (seed=inf). While the Single-Agent setup can also achieve high rewards, the training-time is almost 10 times higher. We further show that setups that have been trained on a single environment only return low cumulative rewards on unseen environments (seed=inf). While the Multi-Agent setup, trained on multiple environments (seed=inf) with an auto-curricula (AC), yields lower mean rewards on the seed 0 environment, it outperforms the other setups - not trained without further environments and auto-curricula - by a wide margin. Additional data on training (\ref{appendix:training-data}) and inference (\ref{appendix:inference-data-2}), including how communication helps our approach to achieve higher performance and cumulative rewards, can be found in the Appendix.

\begin{figure}[!h]
\includegraphics[width=\linewidth]{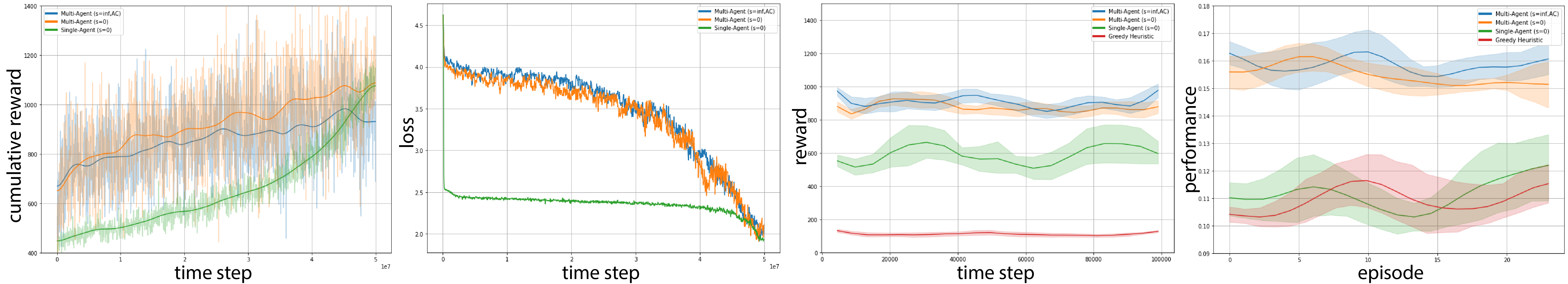} 
\caption{First and second diagram: Training over 5e7 total time steps: cumulative rewards and loss. Third and fourth diagram: Inference on various environments with difficulty level 8: reward vs time step and performance vs episode.}
\label{fig:results}
\end{figure}

\section{Discussion and Future Work}

There are three interesting directions to develop our work further: Firstly, we define neighbourhoods as the three nearest lookout towers, using Euclidean distance. There are further strategies to define neighbourhoods, such as Breath-First-Search (BFS) \citep{burkhardt_optimal_2021}. BFS allows to define multi-layered neighbourhoods. Furthermore incoming messages could be pooled and weighted depending on the distance or neighbourhood-layer of the lookout towers the message is coming from. Secondly, instead of rewarding helping with a bonus reward, we could turn this around and slightly weight rewards higher for actions that benefit agents self first. And lastly we could threshold the support amount one lookout tower is able to receive. In our current approach the centre most lookout tower could hold full support of all eight neighbours, which might not be beneficial for the collective but yield temporarily high egoistic rewards.

\newpage
\subsubsection*{Acknowledgments}
We want to thank Serkan Cabi, Panagiotis Tigas for their guidance and Ian Gemp for igniting the wild fire context idea. We also want to thank the reviewing committee for their efforts and critic. Further information, video material and an interactive web-app can be found at: \url{https://ai.philippsiedler.com/iclr2022-gnn-marl-wild-fire-management-resource-distribution/}.

\bibliography{iclr2022_conference}
\bibliographystyle{gmas_iclr2022_conference}

\appendix
\section{Appendix}
\section{Hyperparameters}
\subsection{Single Agent Training Hyperparameters}
\begin{verbatim}
behaviors:
  WT_SA:
    trainer_type: ppo
    hyperparameters:
      batch_size: 128
      buffer_size: 2048
      learning_rate: 0.0003
      beta: 0.01
      epsilon: 0.2
      lambd: 0.95
      num_epoch: 3
      learning_rate_schedule: linear
    network_settings:
      normalize: false
      hidden_units: 512
      num_layers: 2
      vis_encode_type: simple
    reward_signals:
      extrinsic:
        gamma: 0.99
        strength: 1.0
      curiosity:
        gamma: 0.99
        strength: 0.02
        encoding_size: 256
        learning_rate: 0.0003
    keep_checkpoints: 5
    max_steps: 50000000
    time_horizon: 128
    summary_freq: 40500
    threaded: true
\end{verbatim}

\subsection{Multi Agent Training Hyperparameters}
\begin{verbatim}
behaviors:
  WT_MA:
    trainer_type: ppo
    hyperparameters:
      batch_size: 128
      buffer_size: 2048
      learning_rate: 0.0003
      beta: 0.01
      epsilon: 0.2
      lambd: 0.95
      num_epoch: 3
      learning_rate_schedule: linear
    network_settings:
      normalize: false
      hidden_units: 512
      num_layers: 2
      vis_encode_type: simple
    reward_signals:
      extrinsic:
        gamma: 0.99
        strength: 1.0
      curiosity:
        gamma: 0.99
        strength: 0.02
        encoding_size: 256
        learning_rate: 0.0003
    keep_checkpoints: 5
    max_steps: 5000000
    time_horizon: 128
    summary_freq: 24300
    threaded: true
\end{verbatim}

\subsection{Multi Agent Auto Curriculum Training Hyperparameters}
\label{appendix:multi-agent-auto-curricula}
\begin{verbatim}
behaviors:
  WT_MA:
    trainer_type: ppo
    hyperparameters:
      batch_size: 128
      buffer_size: 2048
      learning_rate: 0.0003
      beta: 0.01
      epsilon: 0.2
      lambd: 0.95
      num_epoch: 3
      learning_rate_schedule: linear
    network_settings:
      normalize: false
      hidden_units: 512
      num_layers: 2
      vis_encode_type: simple
    reward_signals:
      extrinsic:
        gamma: 0.99
        strength: 1.0
      curiosity:
        gamma: 0.99
        strength: 0.02
        encoding_size: 256
        learning_rate: 0.0003
    keep_checkpoints: 5
    max_steps: 100000000
    time_horizon: 128
    summary_freq: 40500
    threaded: true

environment_parameters:
  difficulty:
    curriculum:
      - name: Lesson1
        completion_criteria:
          measure: reward
          behavior: WT_MA
          signal_smoothing: true
          min_lesson_length: 100
          threshold: 900
        value: 1
      - name: Lesson2
        completion_criteria:
          measure: reward
          behavior: WT_MA
          signal_smoothing: true
          min_lesson_length: 100
          threshold: 950
        value: 2
      - name: Lesson3
        completion_criteria:
          measure: reward
          behavior: WT_MA
          signal_smoothing: true
          min_lesson_length: 100
          threshold: 1000
        value: 3
      - name: Lesson4
        completion_criteria:
          measure: reward
          behavior: WT_MA
          signal_smoothing: true
          min_lesson_length: 100
          threshold: 1050
        value: 4
      - name: Lesson5
        completion_criteria:
          measure: reward
          behavior: WT_MA
          signal_smoothing: true
          min_lesson_length: 100
          threshold: 1100
        value: 5
      - name: Lesson6
        completion_criteria:
          measure: reward
          behavior: WT_MA
          signal_smoothing: true
          min_lesson_length: 100
          threshold: 1150
        value: 6
      - name: Lesson7
        completion_criteria:
          measure: reward
          behavior: WT_MA
          signal_smoothing: true
          min_lesson_length: 100
          threshold: 1200
        value: 7
      - name: Lesson8
        completion_criteria:
          measure: reward
          behavior: WT_MA
          signal_smoothing: true
          min_lesson_length: 100
          threshold: 1250
        value: 8
      - name: Lesson9
        completion_criteria:
          measure: reward
          behavior: WT_MA
          signal_smoothing: true
          min_lesson_length: 100
          threshold: 1300
        value: 9
      - name: Lesson10
        value: 10
\end{verbatim}

\newpage
\subsection{Hyperparameter Description}

\begin{table}[h]
  \begin{tabular}{p{0.3\textwidth}p{0.3\textwidth}p{0.3\textwidth}}
    \toprule
    Hyperparameter & Typical Range & Description\\
    \midrule
    Gamma & $0.8-0.995$ & discount factor for future rewards\\
    Lambda & $0.9-0.95$ & used when calculating the Generalized Advantage Estimate (GAE)\\
    Buffer Size & $2048-409600$ & how many experiences should be collected before updating the model\\
    Batch Size & $512-5120$ (continuous), $32-512$ (discrete) & number of experiences used for one iteration of a gradient descent update.\\\
    Number of Epochs & $3-10$ & number of passes through the experience buffer during gradient descent\\
    Learning Rate & $1e-5-1e-3$ & strength of each gradient descent update step\\
    Time Horizon & $32-2048$ & number of steps of experience to collect per-agent before adding it to the experience buffer\\
    Max Steps & $5e5-1e7$ & number of steps of the simulation (multiplied by frame-skip) during the training process\\
    Beta & $1e-4-1e-2$ & strength of the entropy regularization, which makes the policy "more random"\\
    Epsilon & $0.1-0.3$ & acceptable threshold of divergence between the old and new policies during gradient descent updating\\
    Normalize & $true/false$ & weather normalization is applied to the vector observation inputs\\
    Number of Layers & $1-3$ & number of hidden layers present after the observation input\\
    Hidden Units & $32-512$ & number of units in each fully connected layer of the neural network\\
    \midrule
    Intrinsic Curiosity Module\\
    \midrule
    Curiosity Encoding Size & $64-256$ & size of hidden layer used to encode the observations within the intrinsic curiosity module\\
    Curiosity Strength & $0.1-0.001$ & magnitude of the intrinsic reward generated by the intrinsic curiosity module\\
    \bottomrule
  \end{tabular}
\end{table}

\newpage
\section{Pseudocode}

PPO-CLIP pseudocode \citep{openai_proximal_2021, schulman_proximal_2017}:

\begin{algorithm}
	\caption{PPO-Clip}
	
	\begin{algorithmic}[1]
		\item Input: initial policy parameters $\theta_0$, initial value function parameters $\phi_0$
		\For {$k=0,1,2,\ldots$}
		    \State Collect set of trajectories $\mathcal{D}_k$ = \{$\tau_i$\} by running policy $\pi_k = \pi(\theta_k)$ in the environment.
		    \State Compute rewards-to-go $\hat{R_t}$.
		    \State Compute advantage estimates, $\hat{A_t}$ (using any method of     advantage estimation) based on the
		    \State current value function $V_{\phi_k}$
			\State Update the policy by maximizing the PPO-Clip objective:
			\State $\theta_{k+1} = arg\underset{\theta}{max} \frac{1}{|\mathcal{D}_k|T} \sum_{\tau \in \mathcal{D}_k} \sum_{t = 0}^{T} \min \left( \frac{\pi_\theta(a_t|s_t)}{\pi_{\theta_k}(a_t|s_t)}A^{\pi_{\theta_k}}(s_t, a_t), g(\epsilon, A^{\pi_{\theta_k}}(s_t, a_t)) \right)$,
			\State typically via stochastic gradient ascent with Adam.
			\State Fit value function by regression on mean-squared error:
			\State $\phi_{k+1} = arg\underset{\phi}{min} \frac{1}{|\mathcal{D}_k|T} \sum_{\tau \in \mathcal{D}_k} \sum_{t = 0}^{T} \left( (V_{\phi}(s_t)-\hat{R_t} \right)$
			\State typically via some gradient descent algorithm.
		\EndFor
	\end{algorithmic} 
\end{algorithm}

Simple Multi-Agent PPO pseudocode:

\begin{algorithm}
	\caption{Multi-Agent PPO} 
	\begin{algorithmic}[1]
		\For {$iteration=1,2,\ldots$}
			\For {$actor=1,2,\ldots,N$}
				\State Run policy $\pi_{\theta_{old}}$ in environment for $T$ time steps
				\State Compute advantage estimates $\hat{A}_{1},\ldots,\hat{A}_{T}$
			\EndFor
			\State Optimize surrogate $L$ wrt. $\theta$, with $K$ epochs and minibatch size $M\leq NT$
			\State $\theta_{old}\leftarrow\theta$
		\EndFor
	\end{algorithmic} 
\end{algorithm}

\newpage
\section{Training Data Visualisation}
\label{appendix:training-data}
\begin{figure}[H]
\begin{center}
    \begin{subfigure}{0.47\textwidth}
        \includegraphics[width=1\linewidth]{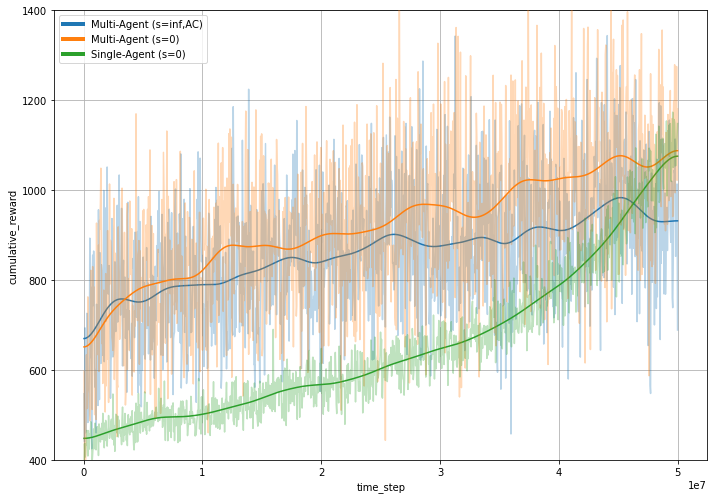}
        \caption{cumulative rewards vs time step}
        \includegraphics[width=1\linewidth]{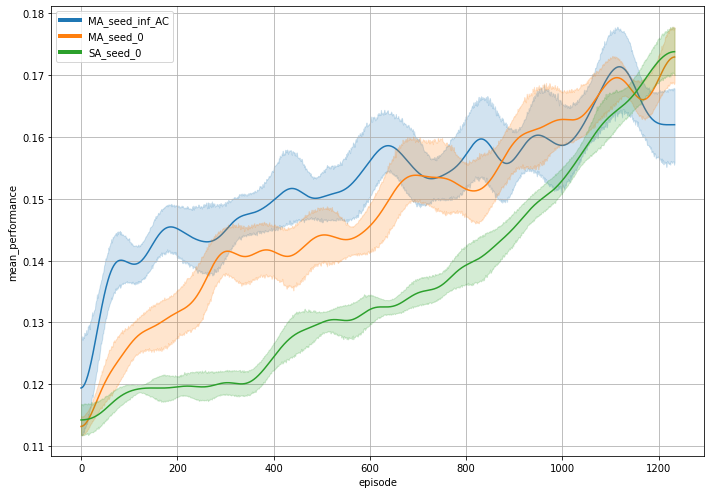}
        \caption{mean performance vs episode}
        \includegraphics[width=1\linewidth]{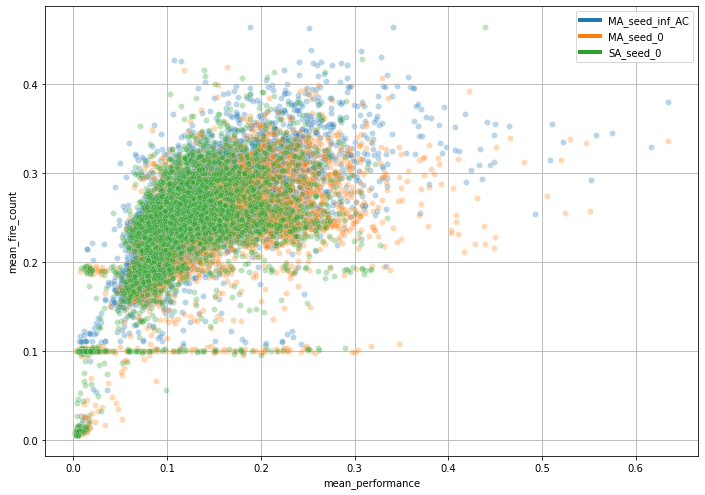}
        \caption{mean fire count vs mean performance}
        \includegraphics[width=1\linewidth]{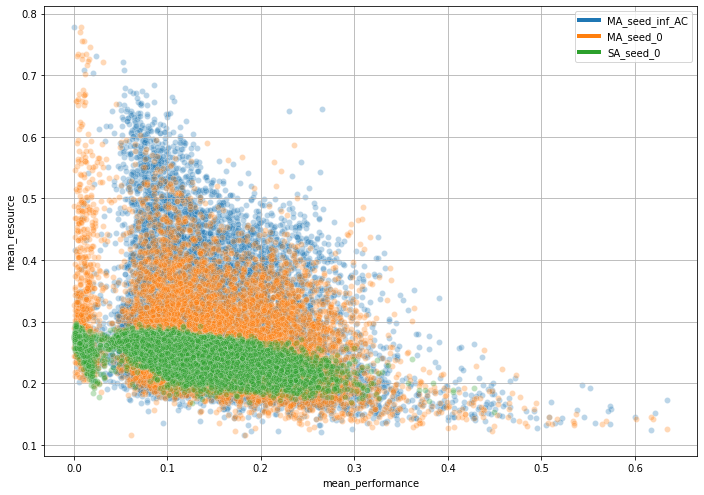}
        \caption{mean resource vs mean performance}
    \end{subfigure}
    \begin{subfigure}{0.47\textwidth}
        \includegraphics[width=1\linewidth]{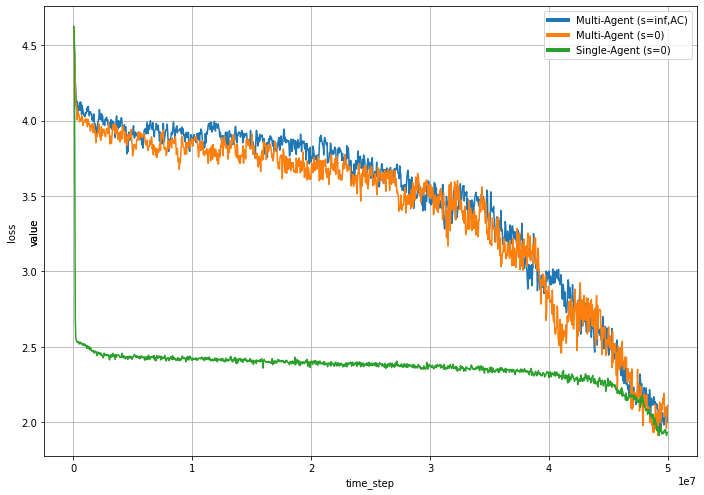}
        \caption{loss vs time step}
        \includegraphics[width=1\linewidth]{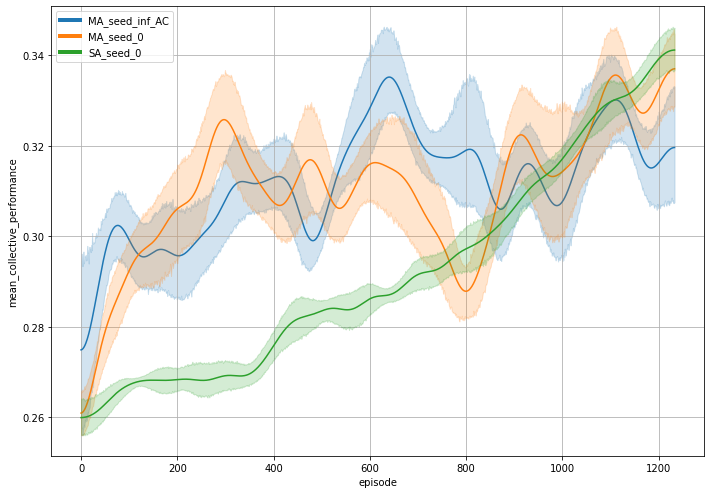}
        \caption{mean collective performance vs episode}
        \includegraphics[width=1\linewidth]{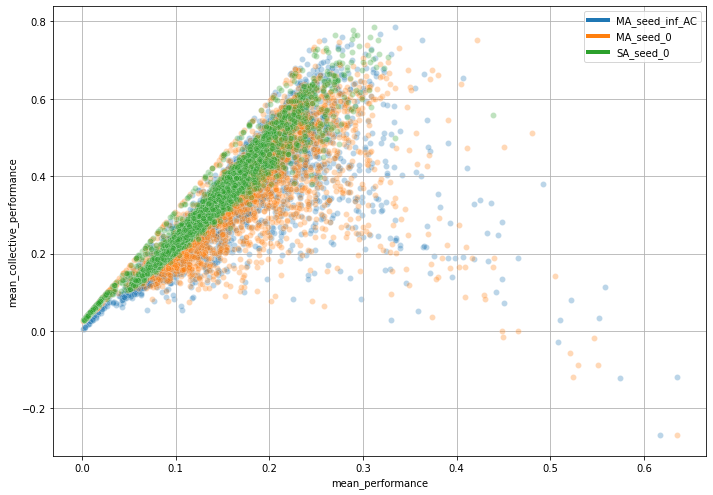}
        \caption{m. collective performance vs mean performance}
        \includegraphics[width=1\linewidth]{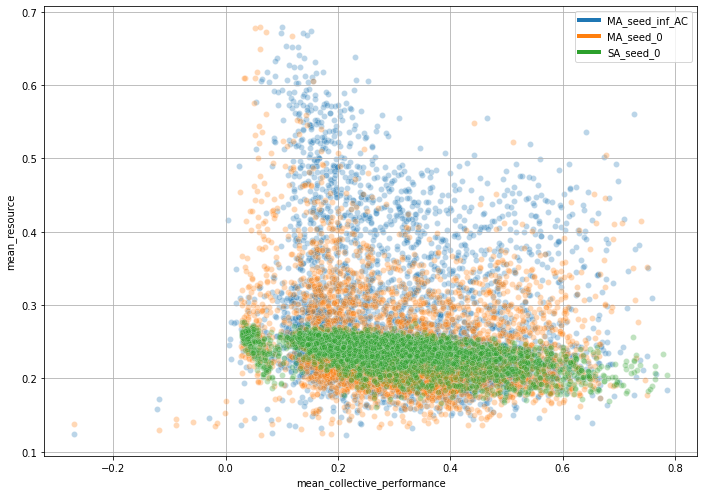}
        \caption{mean resource vs mean collective performance}
    \end{subfigure}
\end{center}
\end{figure}

\newpage

\label{appendix:training-data-1}
\begin{figure}[H]
\begin{center}
    \begin{subfigure}{0.47\textwidth}
        \includegraphics[width=1\linewidth]{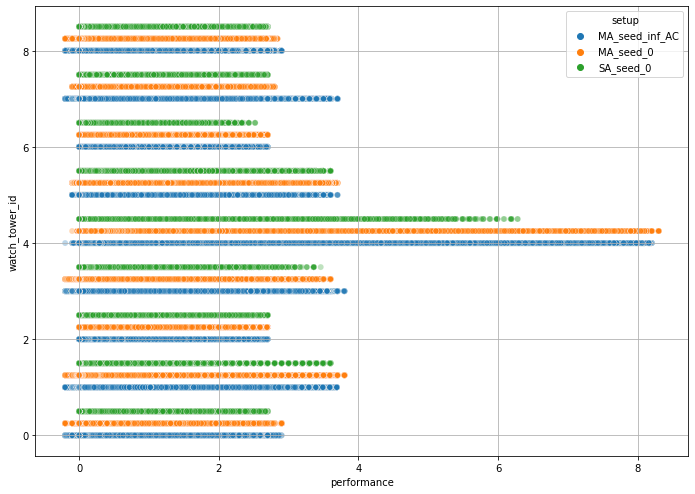}
        \caption{watch tower id vs performance}
        \includegraphics[width=1\linewidth]{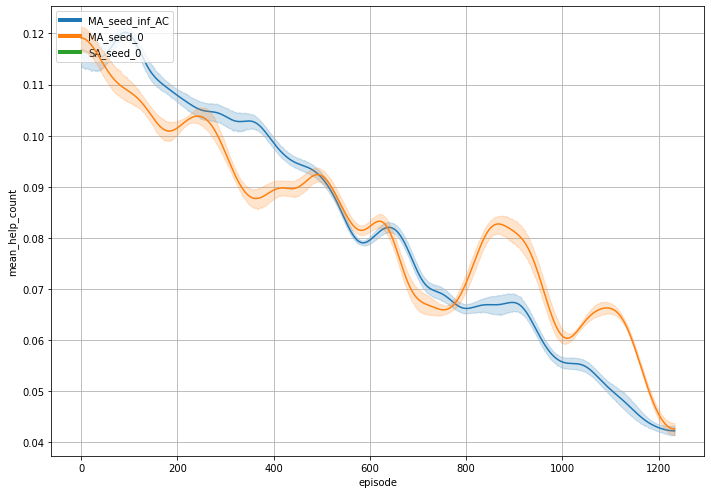}
        \caption{mean help count vs episode}
        \includegraphics[width=1\linewidth]{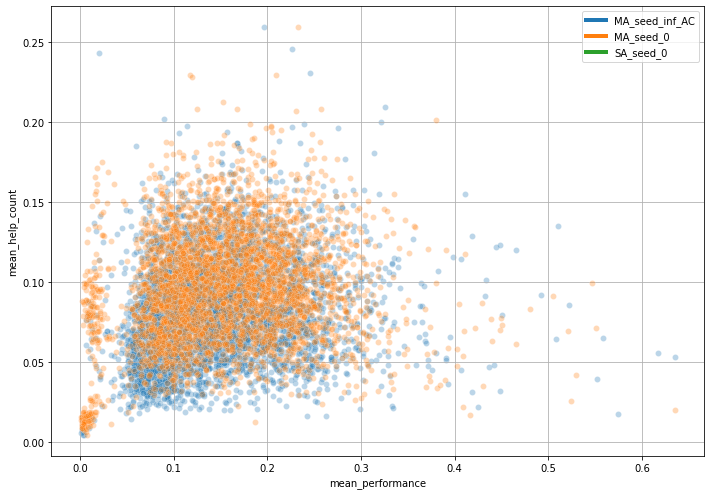}
        \caption{mean help count vs mean performance}
    \end{subfigure}
    \begin{subfigure}{0.47\textwidth}
        \includegraphics[width=1\linewidth]{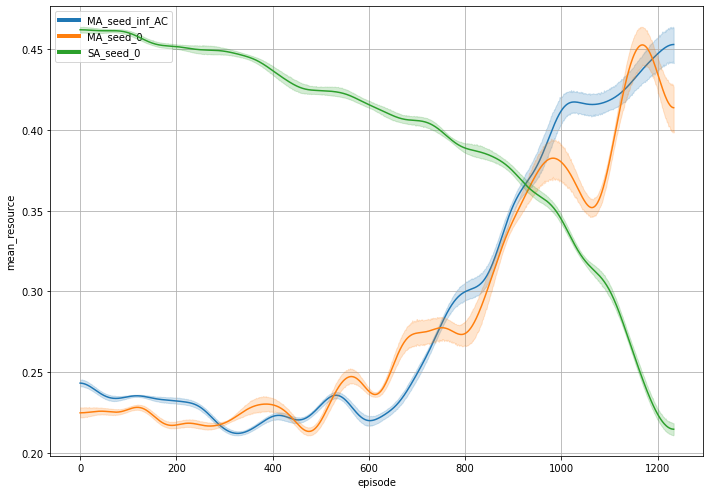}
        \caption{mean resource vs episode}
        \includegraphics[width=1\linewidth]{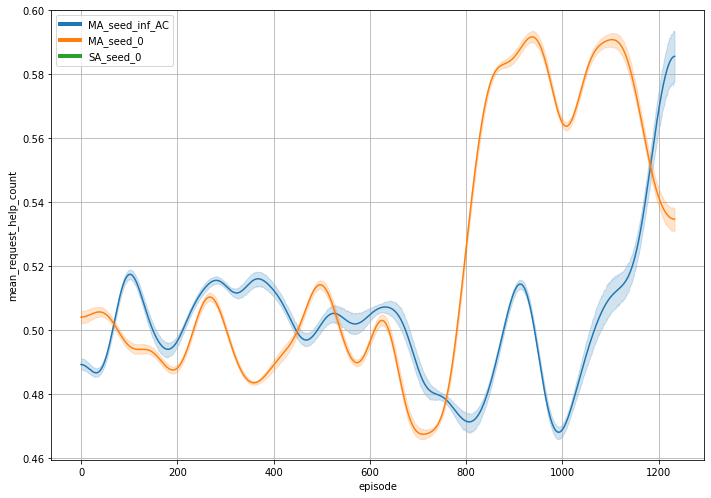}
        \caption{mean request help count vs episode}
        \includegraphics[width=1\linewidth]{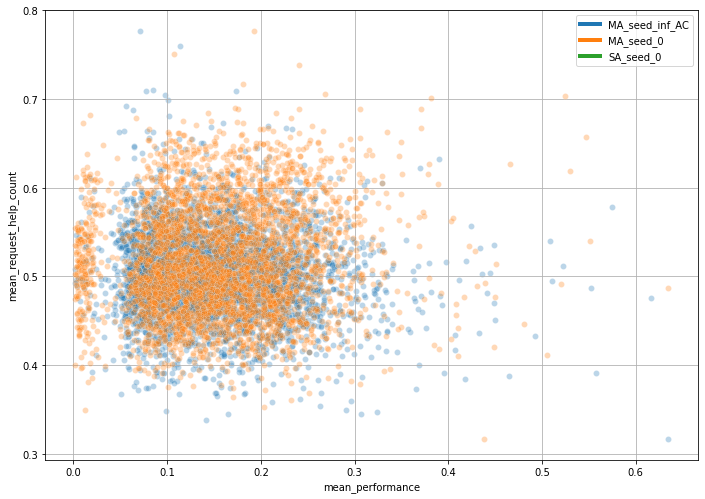}
        \caption{mean request help count vs mean performance}
    \end{subfigure}
\end{center}
\end{figure}

\newpage
\section{Inference Data Visualisation (Left Column: Seed = 0, Right Column: Seed = inf, Auto-Curriculum; starting from figure 10 (b)}
\label{appendix:inference-data-2}
\begin{figure}[H]
\begin{center}
    \begin{subfigure}{0.47\textwidth}
        \includegraphics[width=1\linewidth]{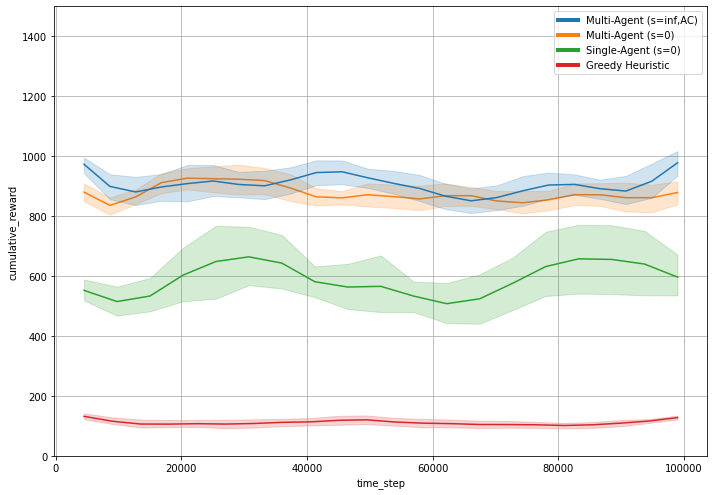}
        \caption{cumulative reward vs time step}
        \includegraphics[width=1\linewidth]{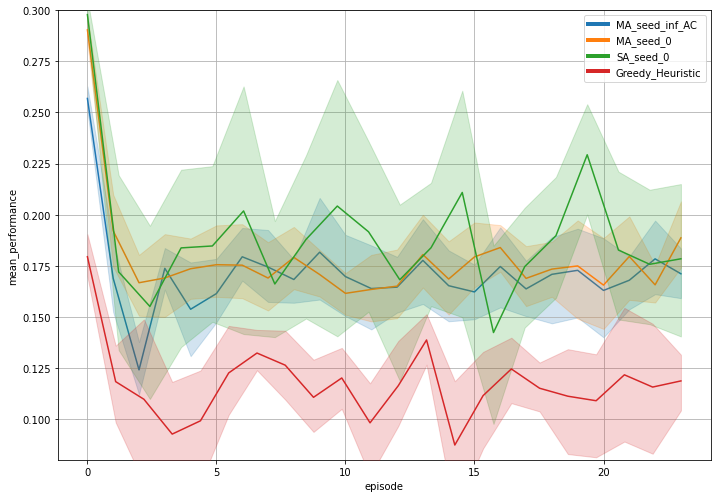}
        \caption{mean performance vs episode}
        \includegraphics[width=1\linewidth]{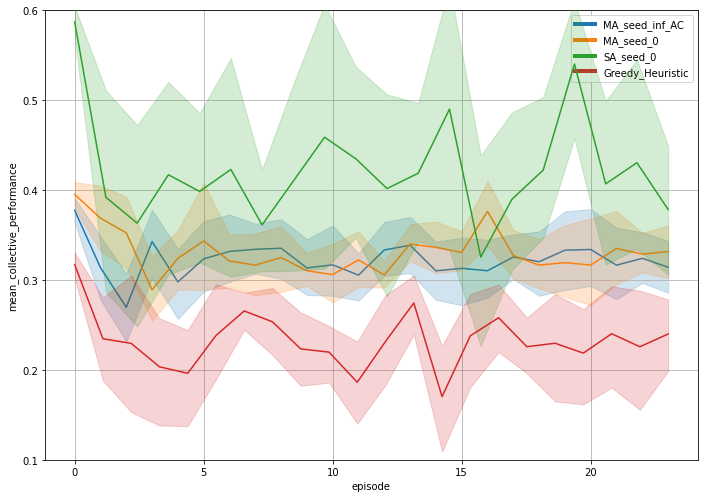}
        \caption{mean collective performance vs episode}
        \caption{Seed = 0}
    \end{subfigure}
    \begin{subfigure}{0.47\textwidth}
    \includegraphics[width=1\linewidth]{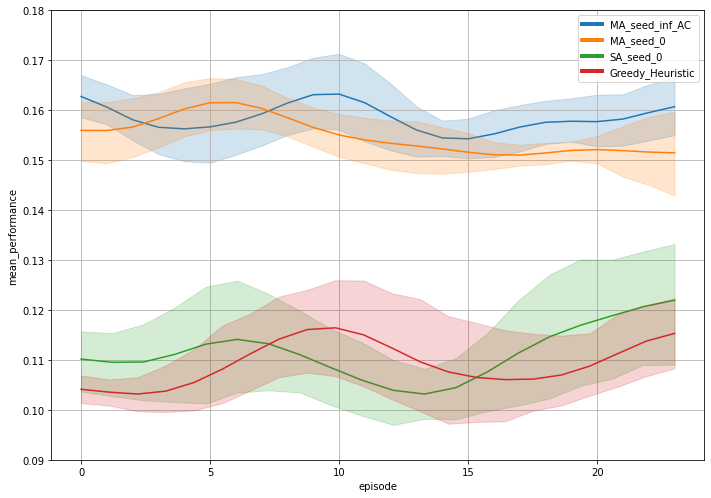}
        \caption{mean performance vs episode}
        \includegraphics[width=1\linewidth]{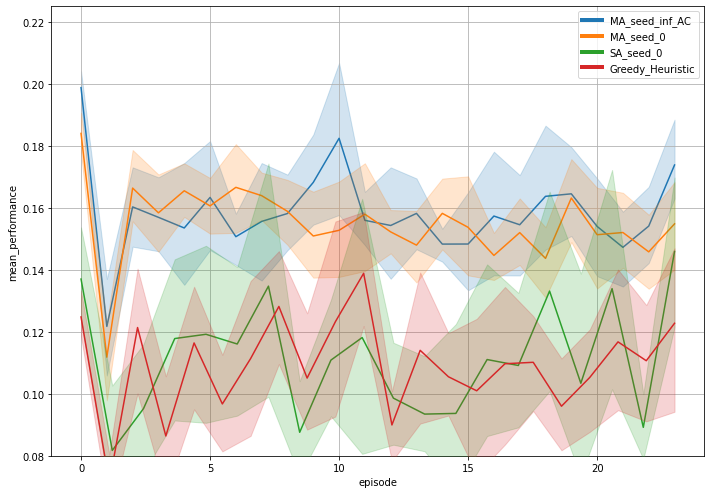}
        \caption{mean performance vs episode}
        \includegraphics[width=1\linewidth]{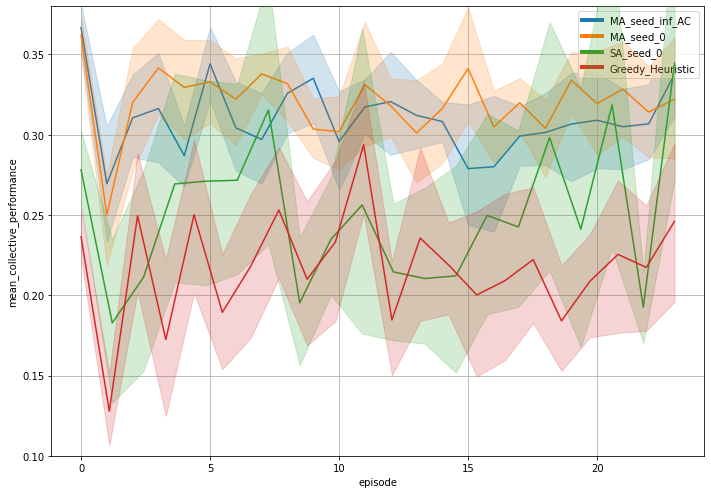}
        \caption{mean collective performance vs episode}
        \caption{Seed = inf, Auto-Curriculum}
    \end{subfigure}
\end{center}
\caption{}
\end{figure}

\newpage

\label{appendix:inference-data-3}
\begin{figure}[H]
\begin{center}
    \begin{subfigure}{0.47\textwidth}
        \includegraphics[width=1\linewidth]{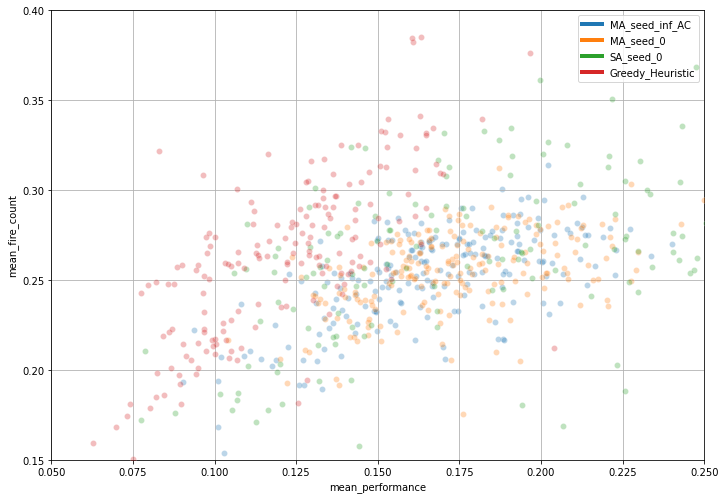}
        \caption{mean fire count vs mean performance}
        \includegraphics[width=1\linewidth]{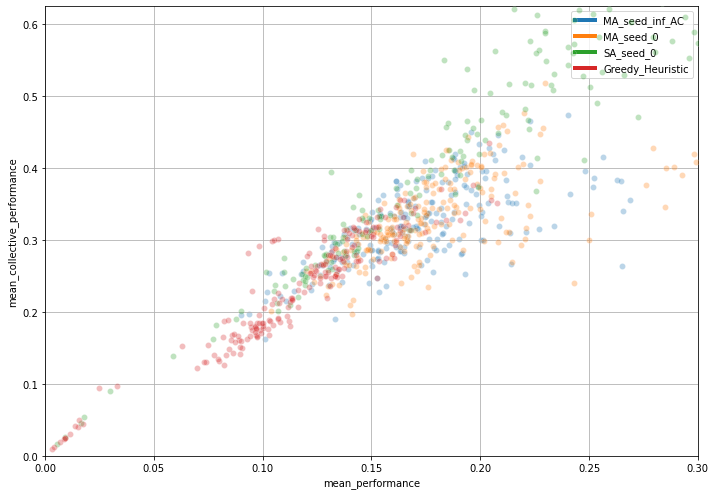}
        \caption{mean collective performance vs mean performance}
        \includegraphics[width=1\linewidth]{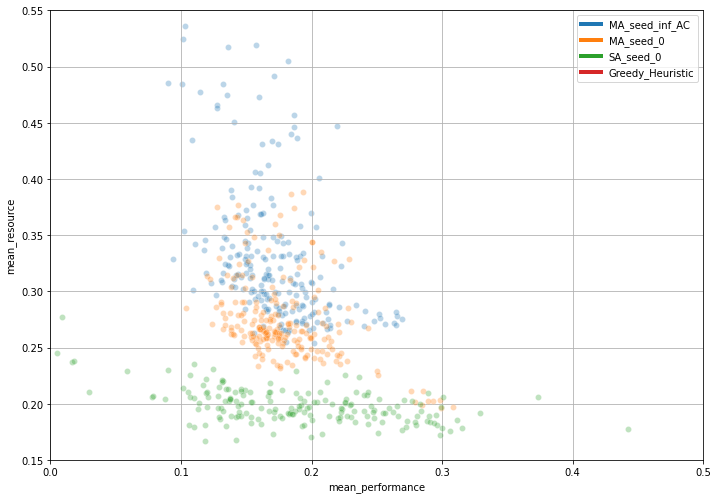}
        \caption{mean resource vs mean performance}
        \includegraphics[width=1\linewidth]{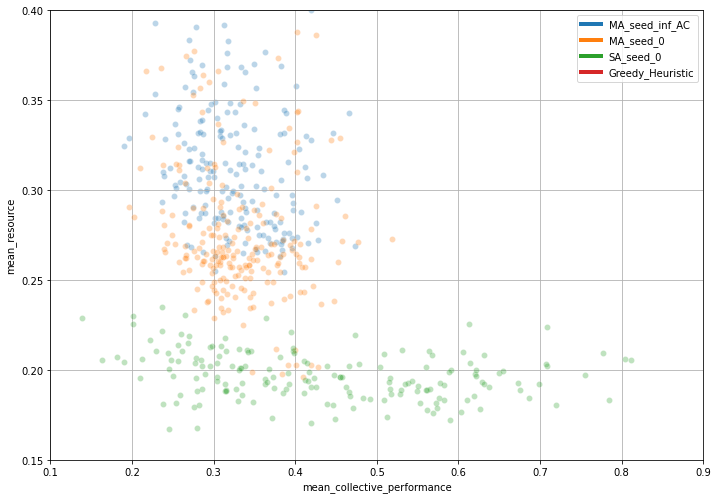}
        \caption{mean resource vs mean collective performance}
        \caption{Seed = 0}
    \end{subfigure}
    \begin{subfigure}{0.47\textwidth}
        \includegraphics[width=1\linewidth]{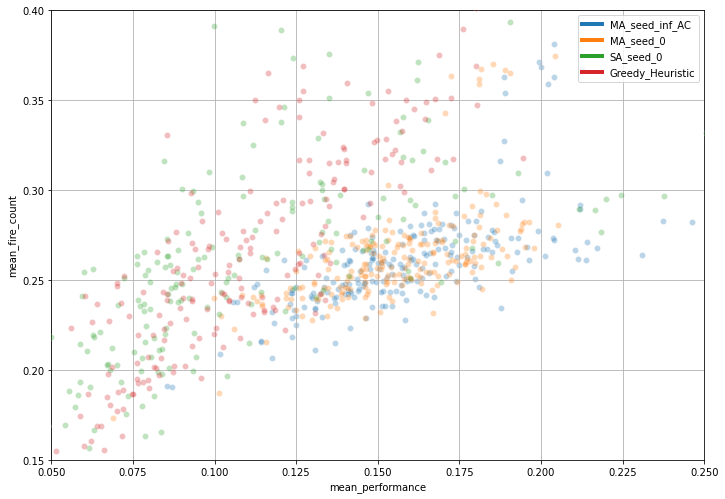}
        \caption{mean fire count vs mean performance}
        \includegraphics[width=1\linewidth]{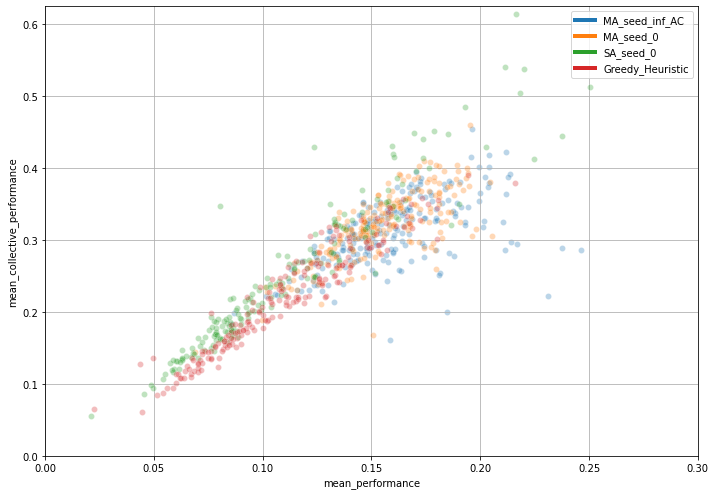}
        \caption{mean collective performance vs mean performance}
        \includegraphics[width=1\linewidth]{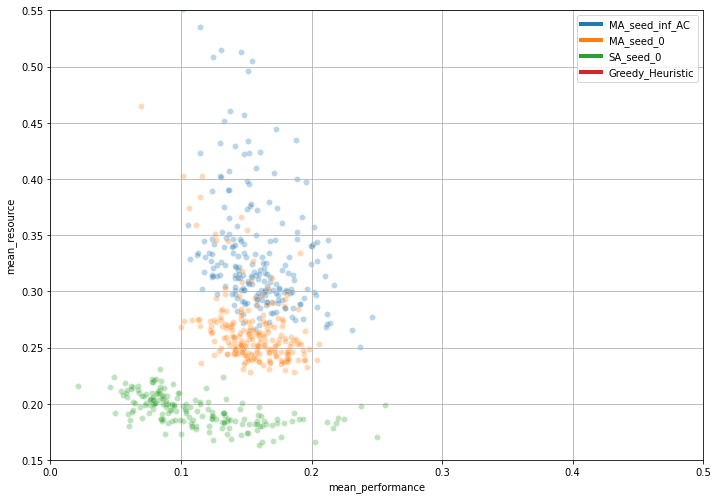}
        \caption{mean resource vs mean performance}
        \includegraphics[width=1\linewidth]{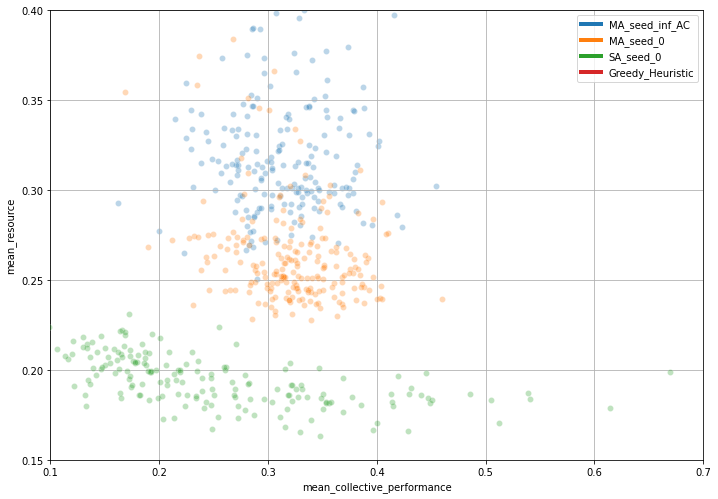}
        \caption{mean resource vs mean collective performance}
        \caption{Seed = inf, Auto-Curriculum}
    \end{subfigure}
\end{center}
\caption{}
\end{figure}

\newpage

\label{appendix:inference-data-4}
\begin{figure}[H]
\begin{center}
    \begin{subfigure}{0.47\textwidth}
        \includegraphics[width=1\linewidth]{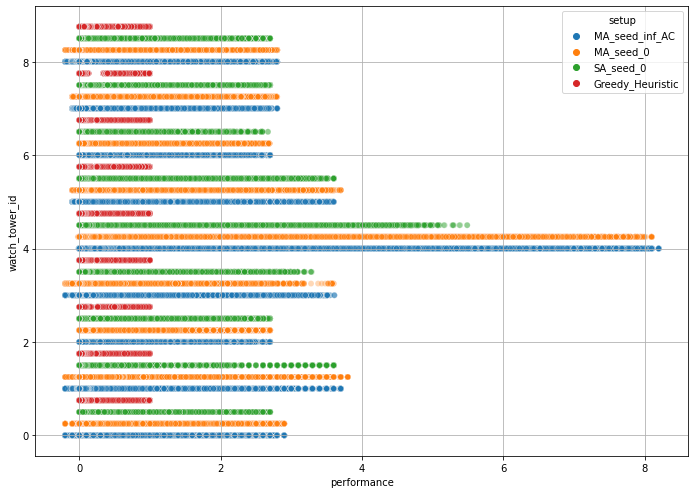}
        \caption{watch tower id vs performance}
        \includegraphics[width=1\linewidth]{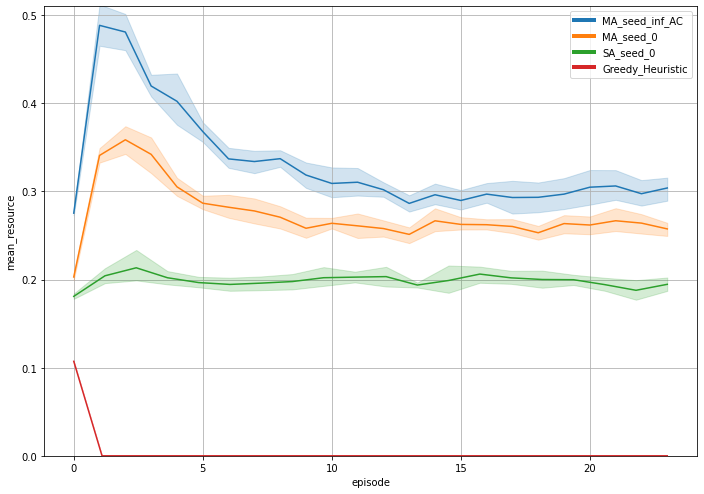}
        \caption{mean resource vs episode}
        \includegraphics[width=1\linewidth]{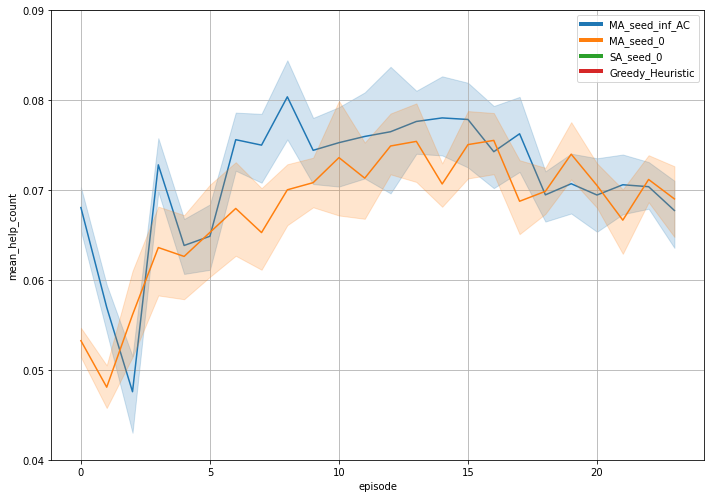}
        \caption{mean help count vs episode}
        \caption{Seed = 0}
    \end{subfigure}
    \begin{subfigure}{0.47\textwidth}
        \includegraphics[width=1\linewidth]{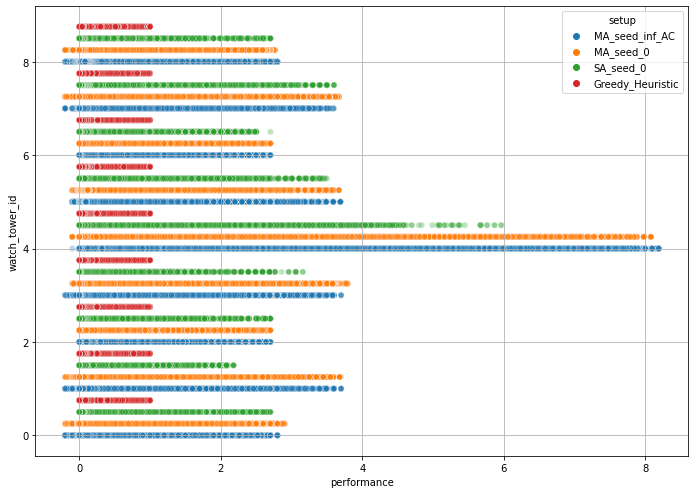}
        \caption{watch tower id vs performance}
        \includegraphics[width=1\linewidth]{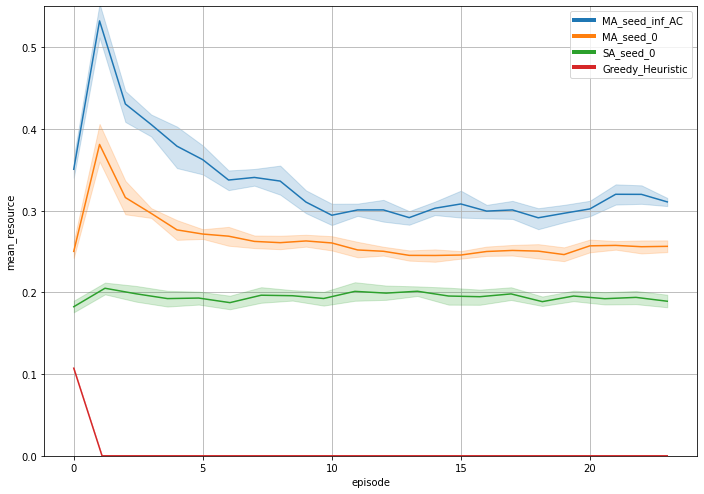}
        \caption{mean resource vs episode}
        \includegraphics[width=1\linewidth]{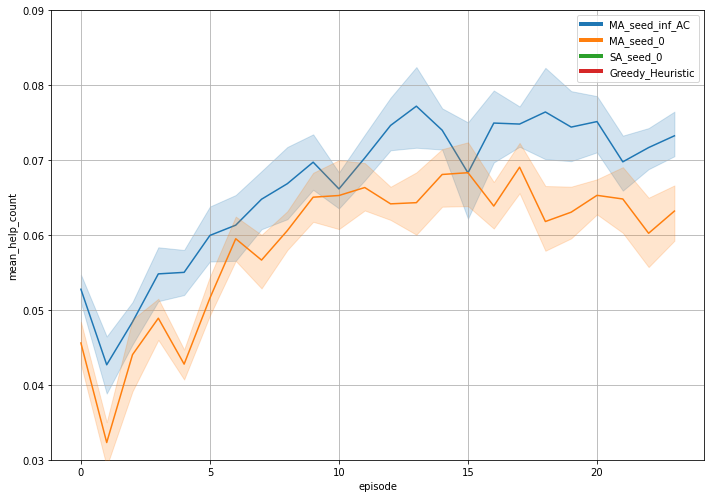}
        \caption{mean help count vs episode}
        \caption{Seed = inf, Auto-Curriculum}
    \end{subfigure}
\end{center}
\caption{}
\end{figure}

\newpage

\label{appendix:inference-data-5}
\begin{figure}[H]
\begin{center}
    \begin{subfigure}{0.47\textwidth}
        \includegraphics[width=1\linewidth]{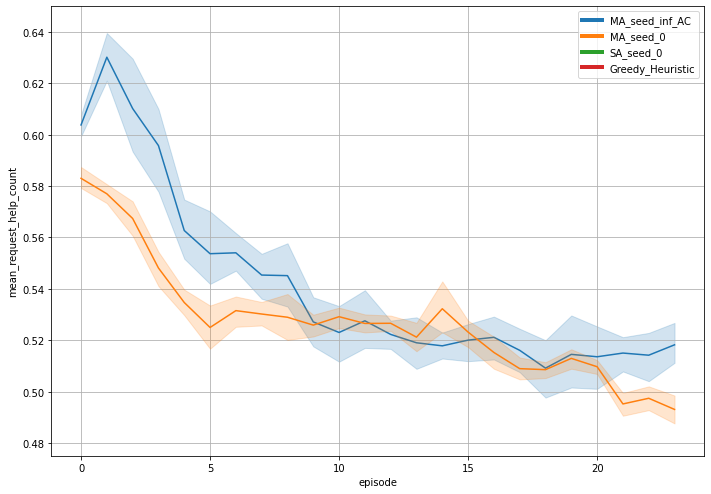}
        \caption{mean help request vs episode}
        \includegraphics[width=1\linewidth]{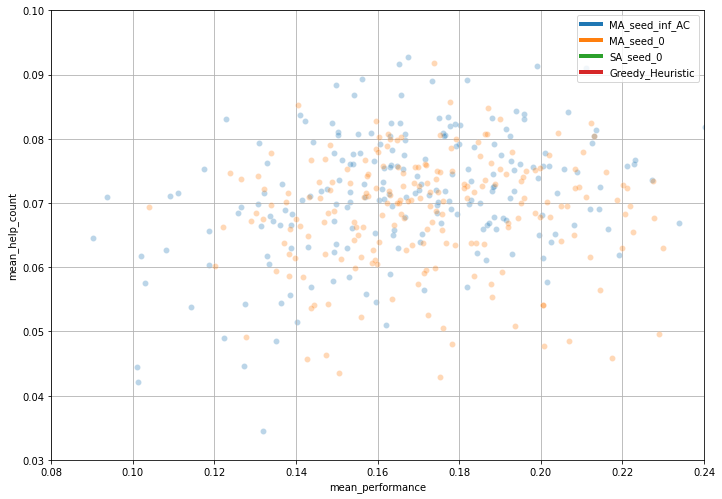}
        \caption{mean help count vs mean performance}
        \includegraphics[width=1\linewidth]{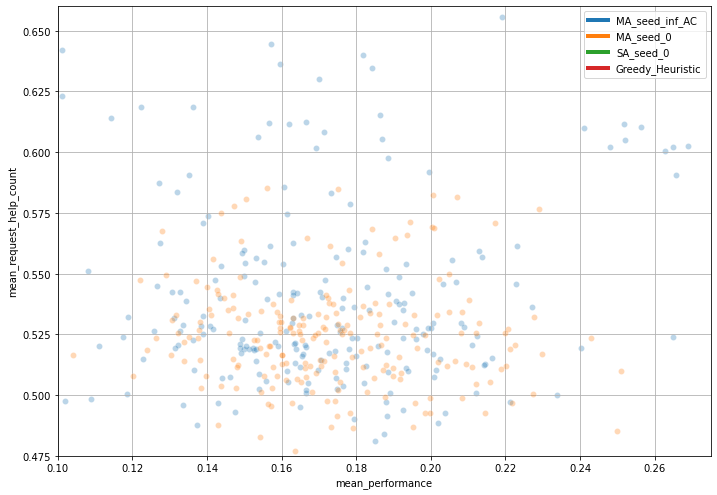}
        \caption{mean request help count vs mean performance}
        \caption{Seed = 0}
    \end{subfigure}
    \begin{subfigure}{0.47\textwidth}
        \includegraphics[width=1\linewidth]{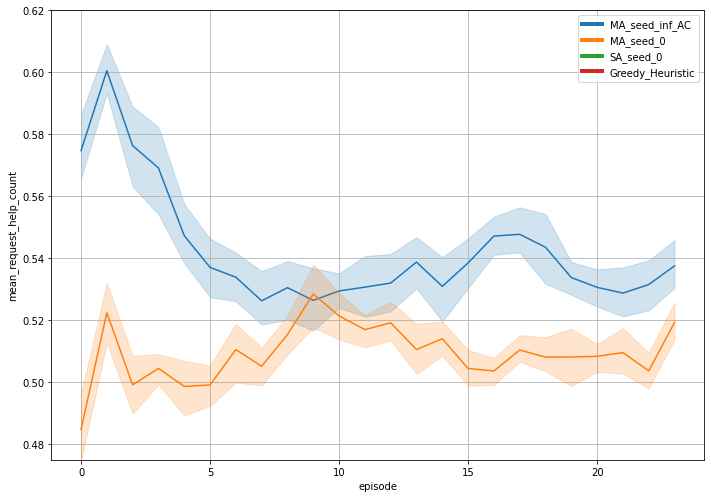}
        \caption{mean help request vs episode}
        \includegraphics[width=1\linewidth]{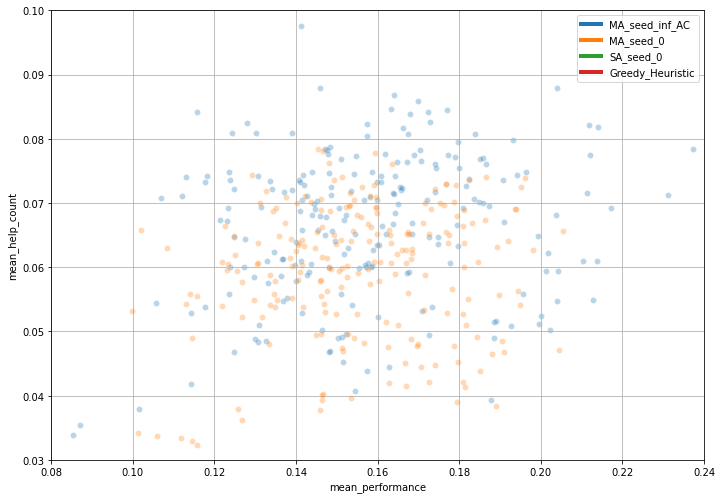}
        \caption{mean help count vs mean performance}
        \includegraphics[width=1\linewidth]{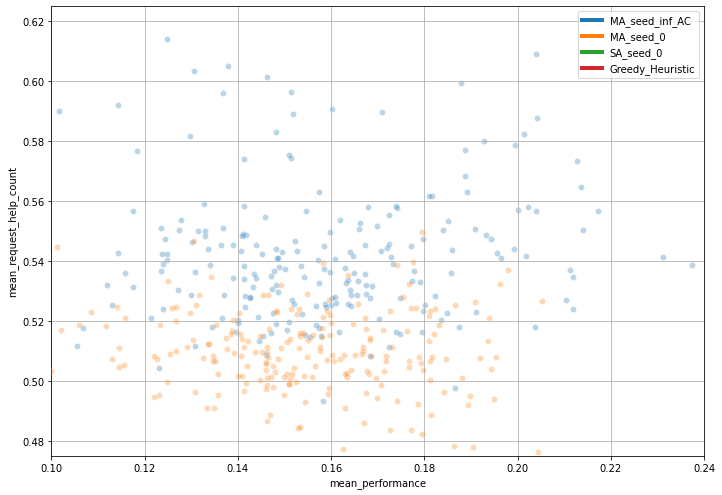}
        \caption{mean request help count vs mean performance}
        \caption{Seed = inf, Auto-Curriculum}
    \end{subfigure}
\end{center}
\caption{}
\end{figure}

\newpage
\section{Environment Scenario Samples: Difficulty VS. Seed Matrix}
\label{appendix:difficulty-vs-seed-matrix}
\begin{figure}[H]
    \includegraphics[width=\textwidth]{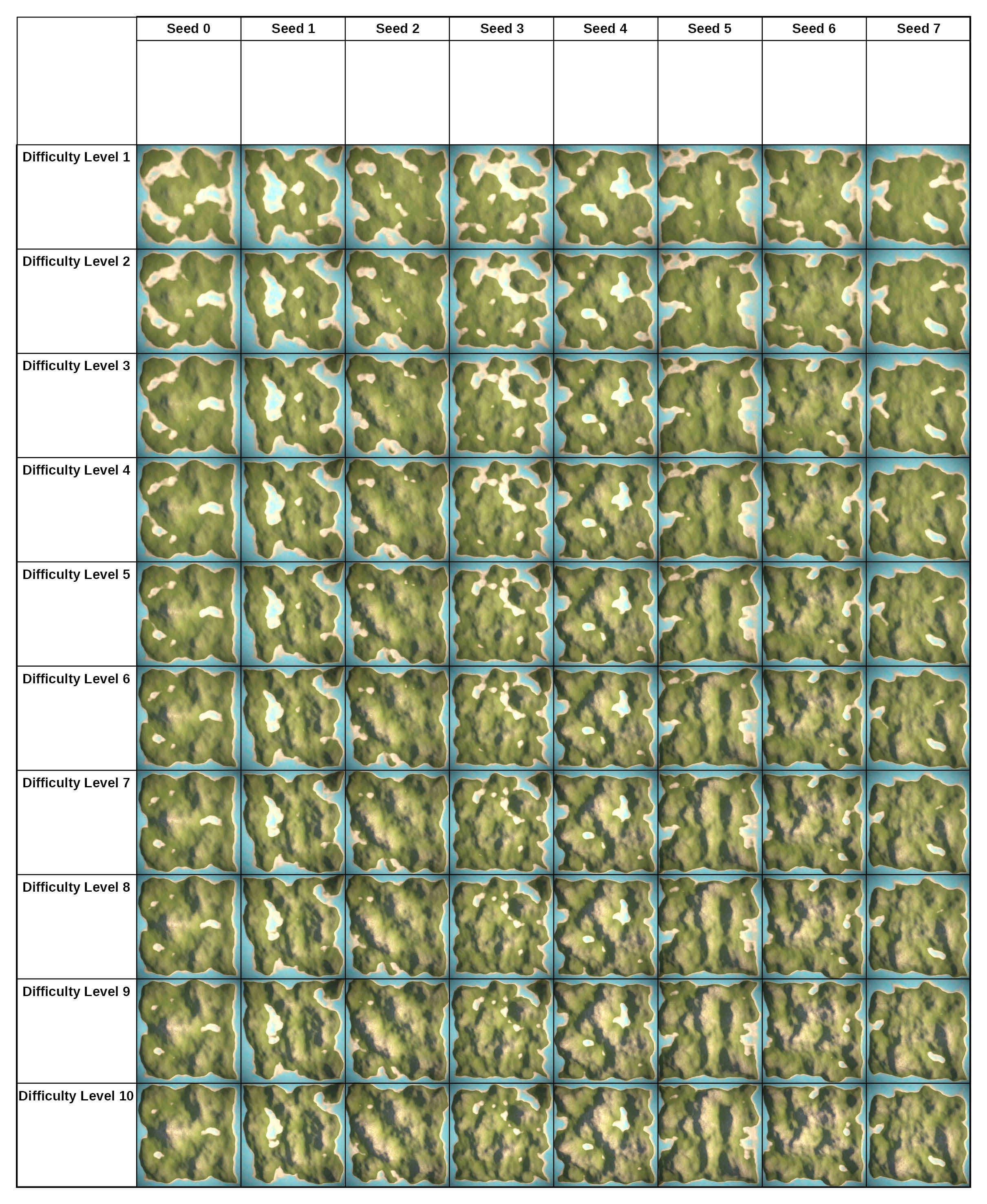}
    \caption{}
\end{figure}

\newpage
\section{Multi-Agent Communication}
\begin{figure}[H]
\includegraphics[width=1\linewidth]{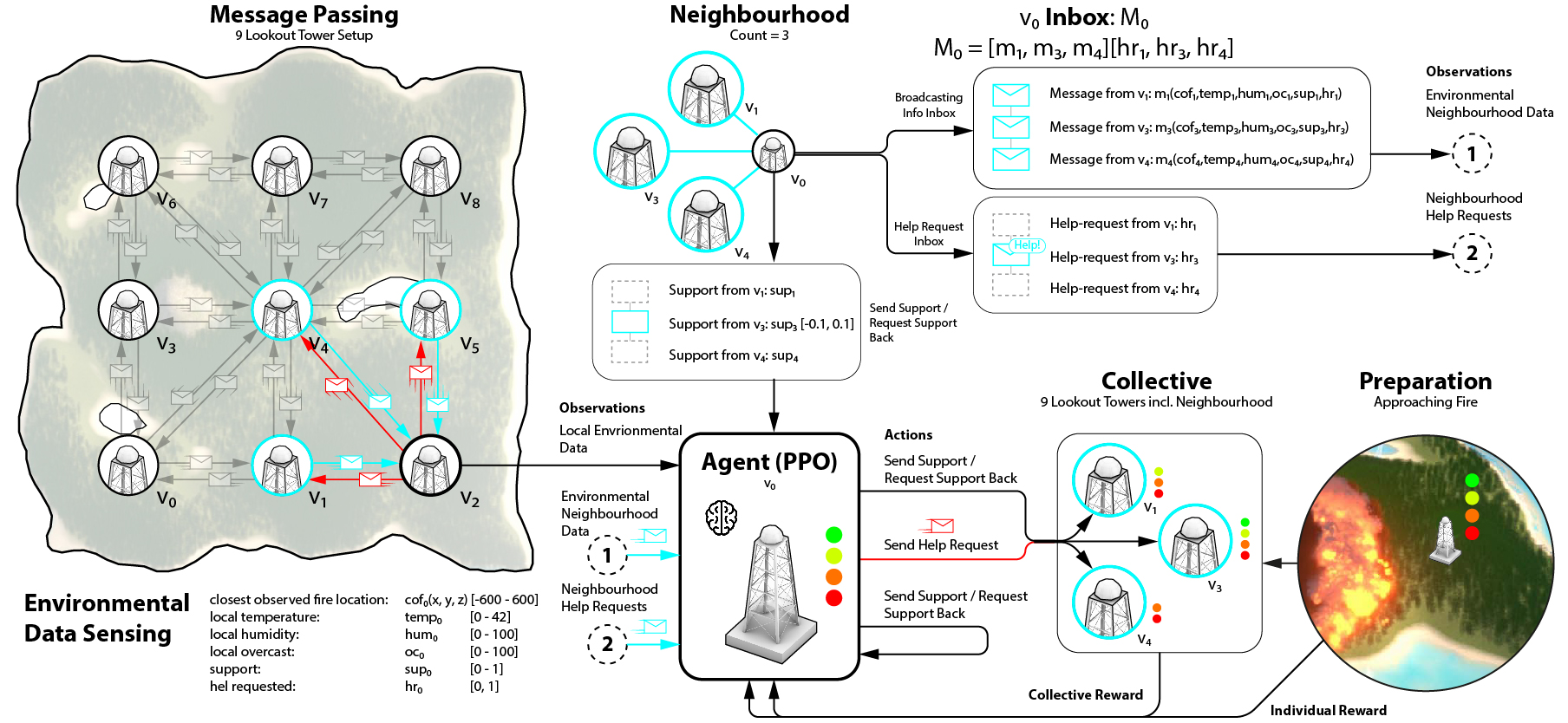}
\caption{}
\label{fig:communication_diagram_large}
\end{figure}

\section{Egoistic Reward Function Diagram}
\begin{figure}[H]
\includegraphics[width=1\linewidth]{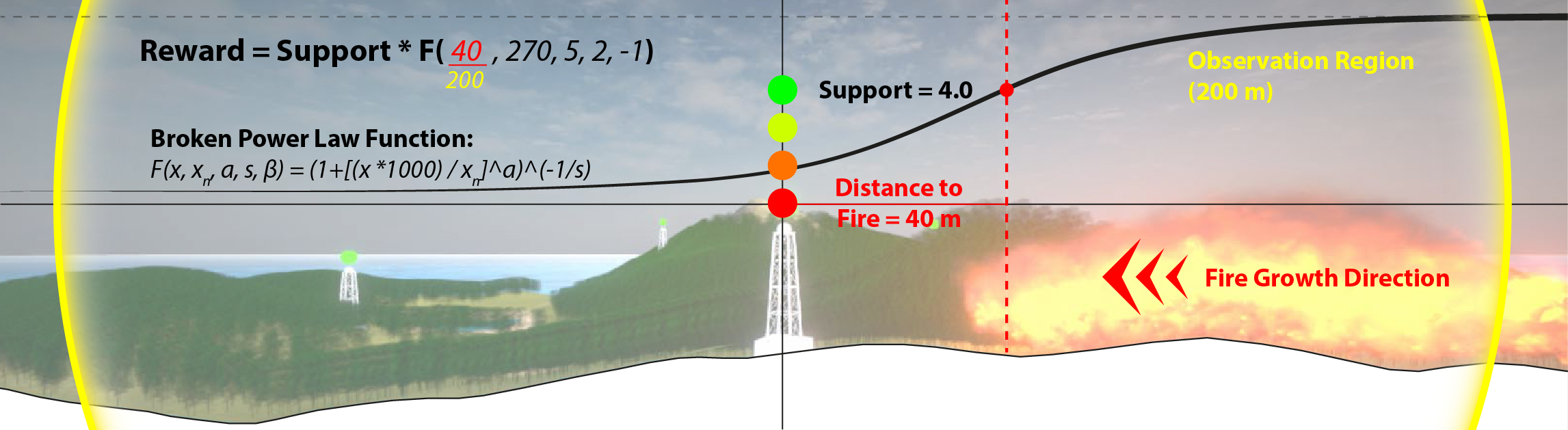}
\caption{}
\label{fig:reward_function_detail}
\end{figure}

\section{Static and Dynamic Environment Features}
\label{appendix:static-and-dynamic-env-1}
\begin{figure}[H]
    \begin{subfigure}{0.495\textwidth}
        \includegraphics[width=1\linewidth]{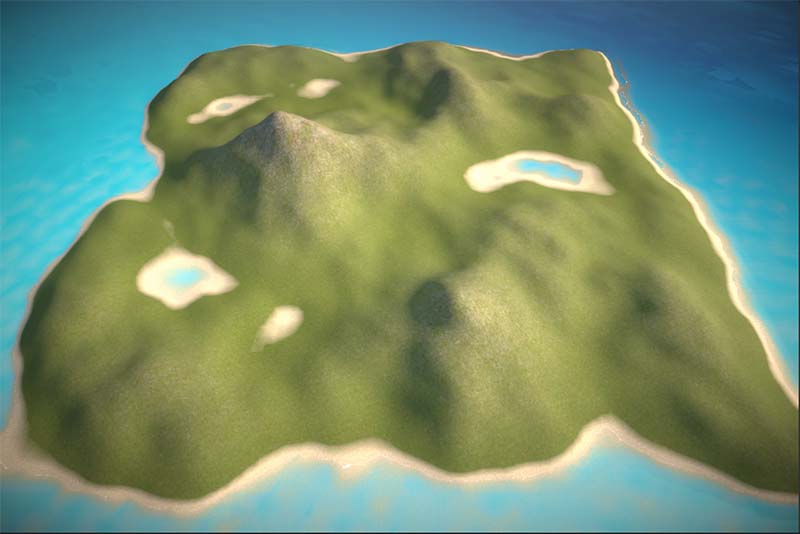}
        \caption{Terrain}
        \caption{Static Features}
    \end{subfigure}
    \begin{subfigure}{0.495\textwidth}
        \includegraphics[width=1\linewidth]{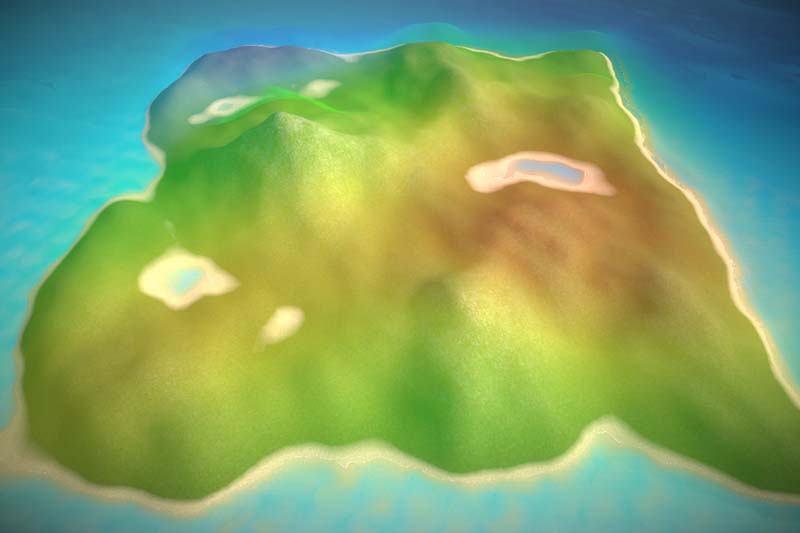}
        \caption{Temperature Heatmap}
        \caption{Dynamic Features}
    \end{subfigure}
\caption{}
\end{figure}

\label{appendix:static-and-dynamic-env-2}
\begin{figure}[H]
    \begin{subfigure}{0.495\textwidth}
        \includegraphics[width=1\linewidth]{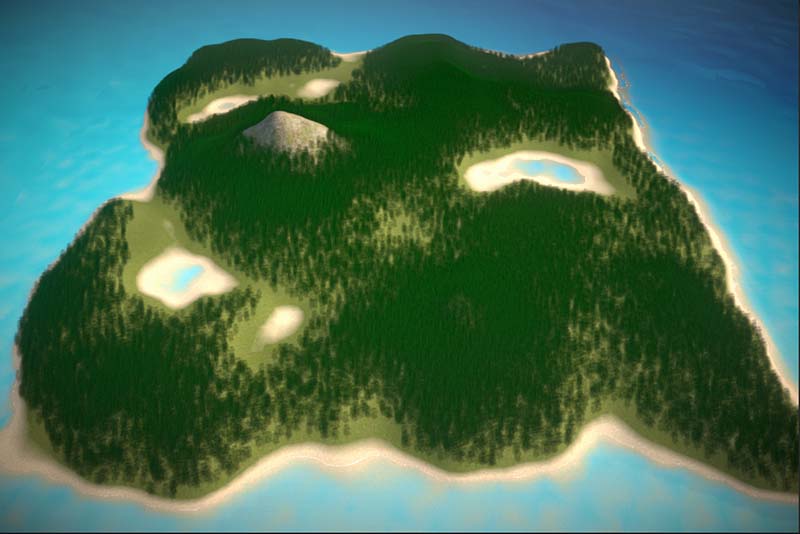}
        \caption{Forest}
        \includegraphics[width=1\linewidth]{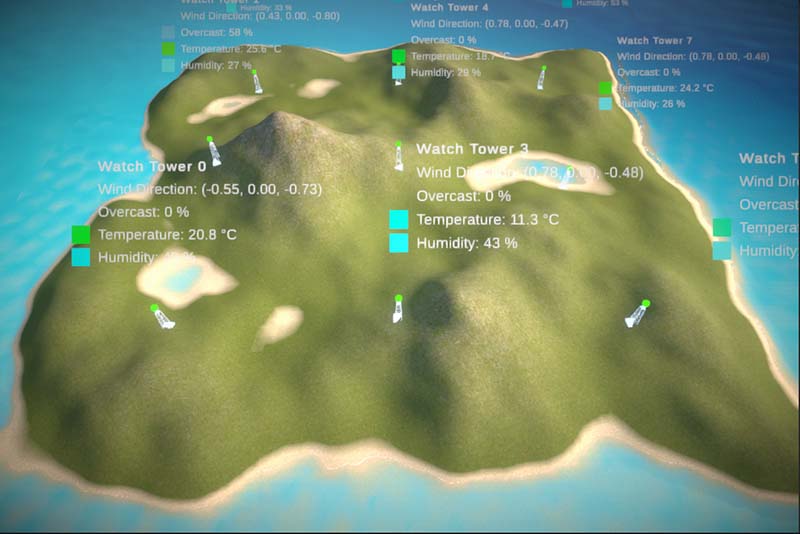}
        \caption{Info Tags}
        \includegraphics[width=1\linewidth]{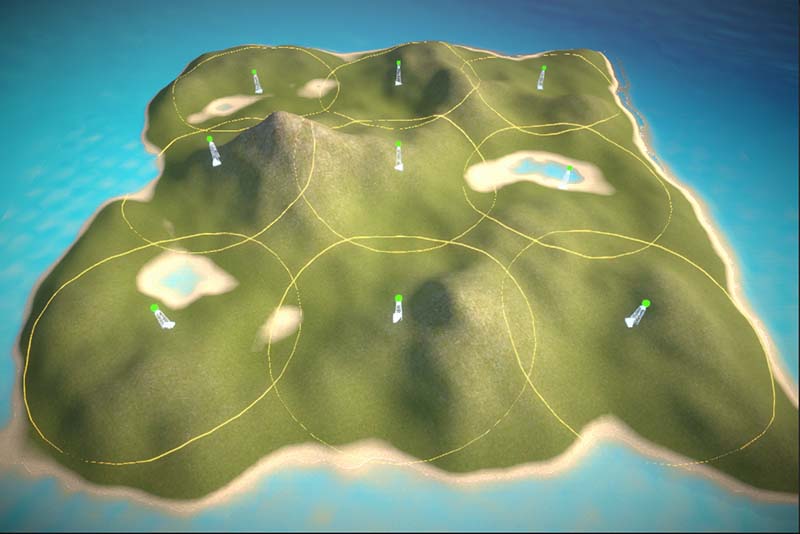}
        \caption{Lookout Tower Observation Region}
        \includegraphics[width=1\linewidth]{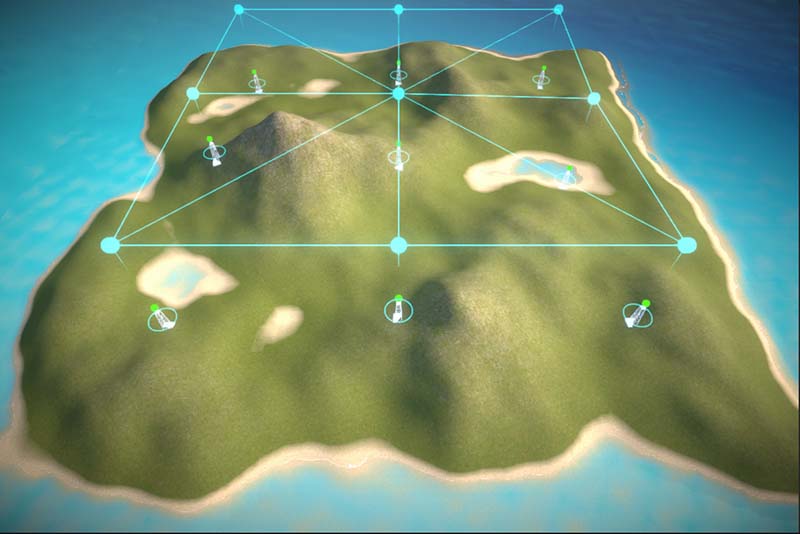}
        \caption{Neighbourhood Network}
        \caption{Static Features}
    \end{subfigure}
    \begin{subfigure}{0.495\textwidth}
        \includegraphics[width=1\linewidth]{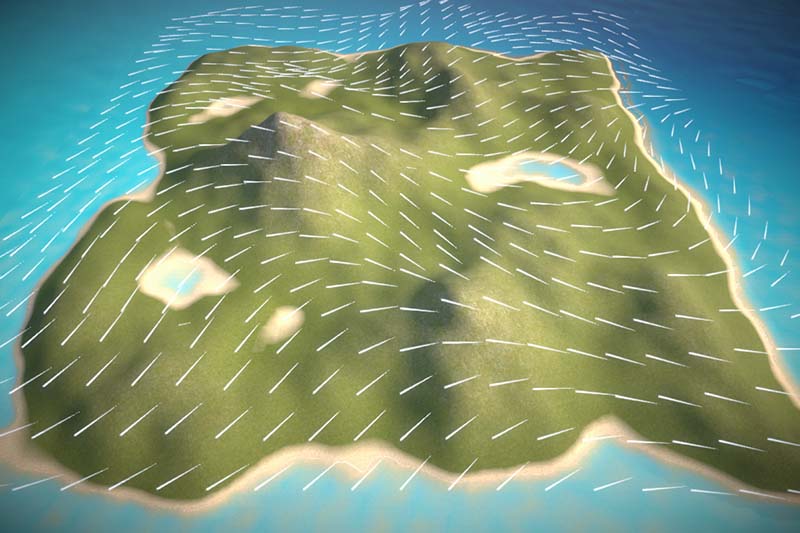}
        \caption{Wind Field}
        \includegraphics[width=1\linewidth]{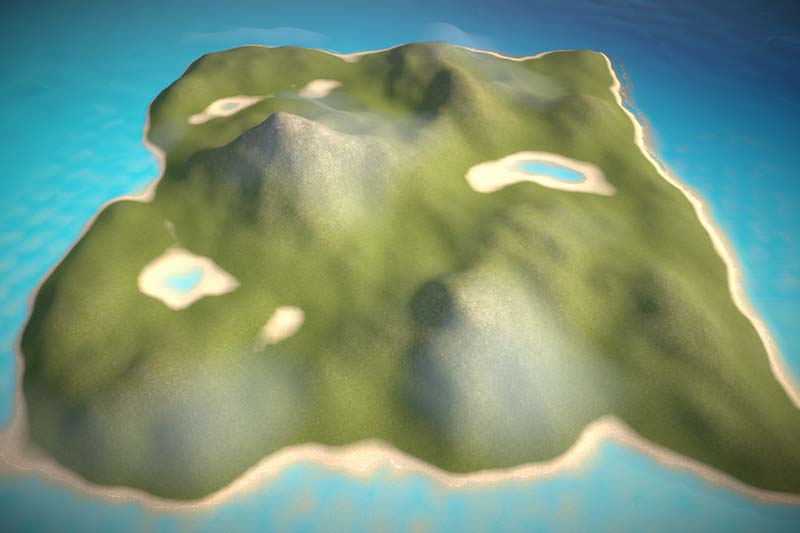}
        \caption{Overcast}
        \includegraphics[width=1\linewidth]{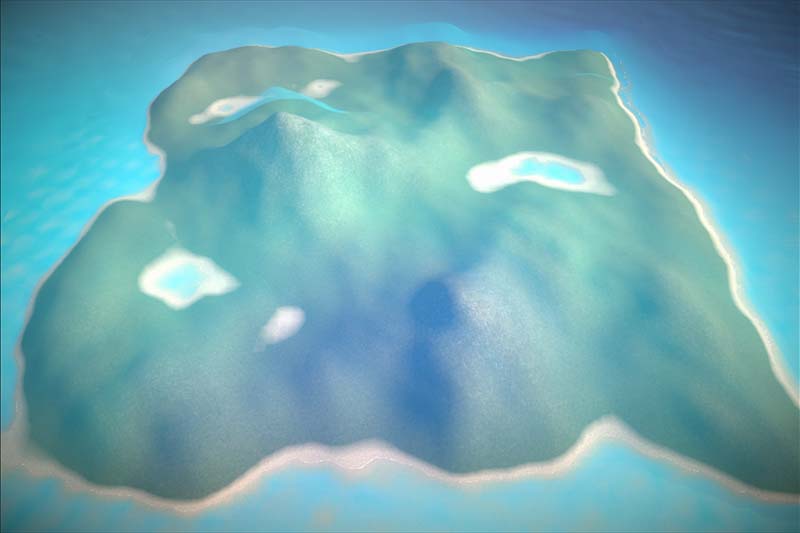}
        \caption{Humidity Heatmap}
        \includegraphics[width=1\linewidth]{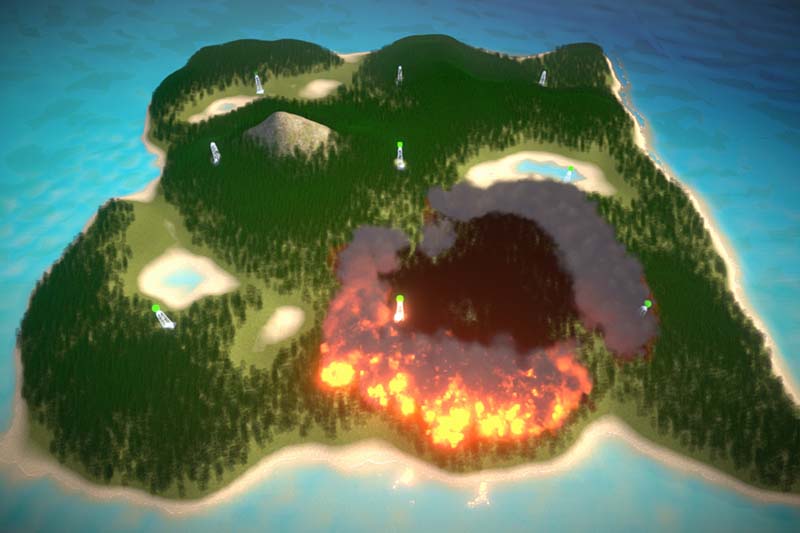}
        \caption{Wild Fire}
        \caption{Dynamic Features}
    \end{subfigure}
\caption{}
\end{figure}

\newpage
\section{Wild Fire Growth Behaviour Frames}
\label{appendix:fire-frames}
\begin{figure}[H]
    \begin{subfigure}{0.495\textwidth}
        \includegraphics[width=1\linewidth]{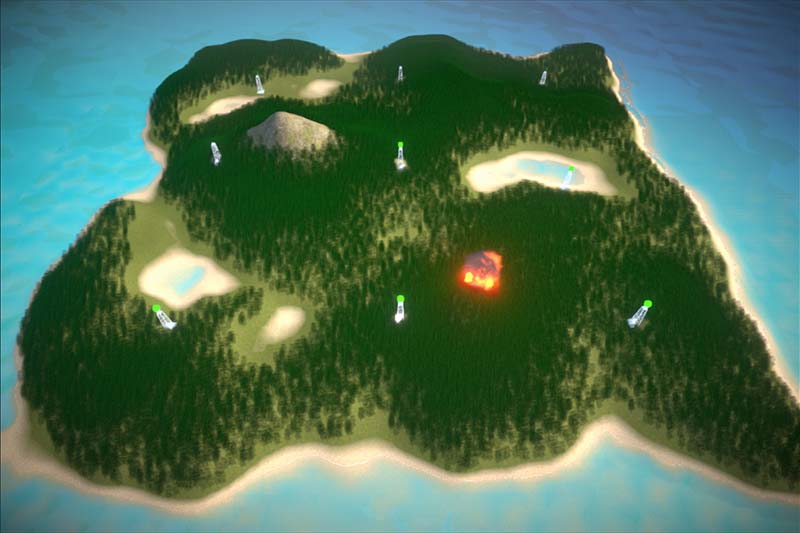}
        \caption{Frame 1}
        \includegraphics[width=1\linewidth]{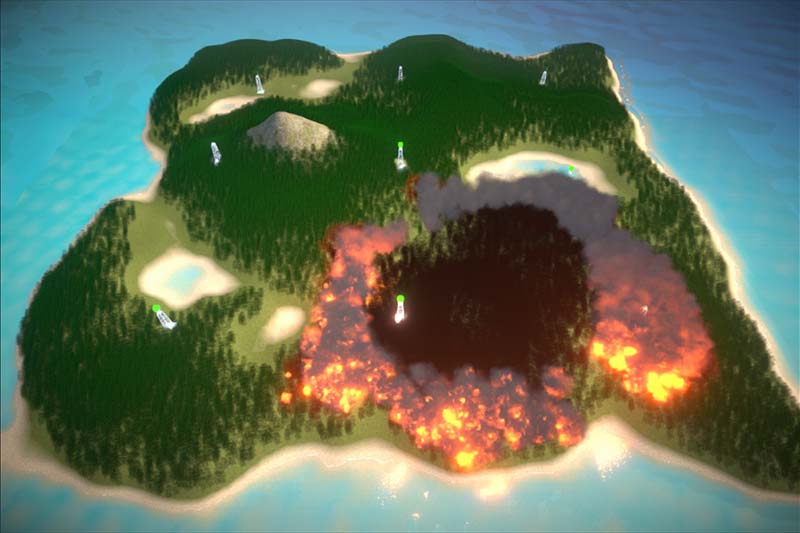}
        \caption{Frame 3}
        \includegraphics[width=1\linewidth]{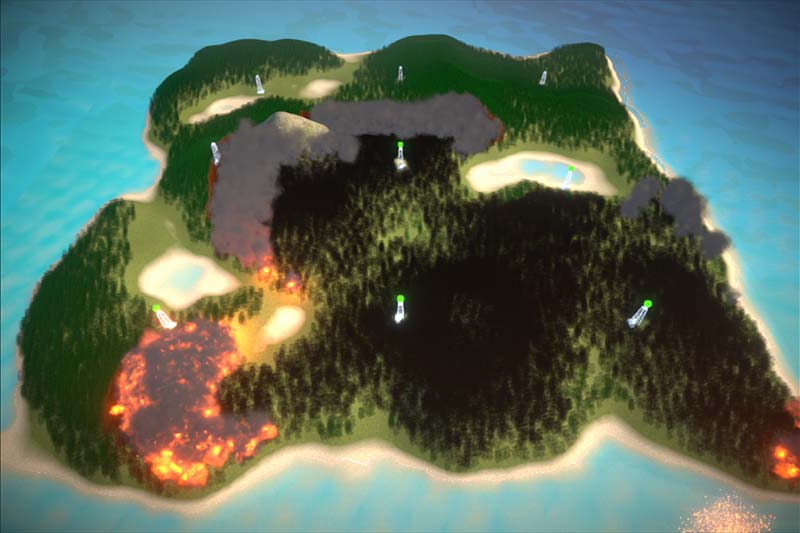}
        \caption{Frame 5}
        \includegraphics[width=1\linewidth]{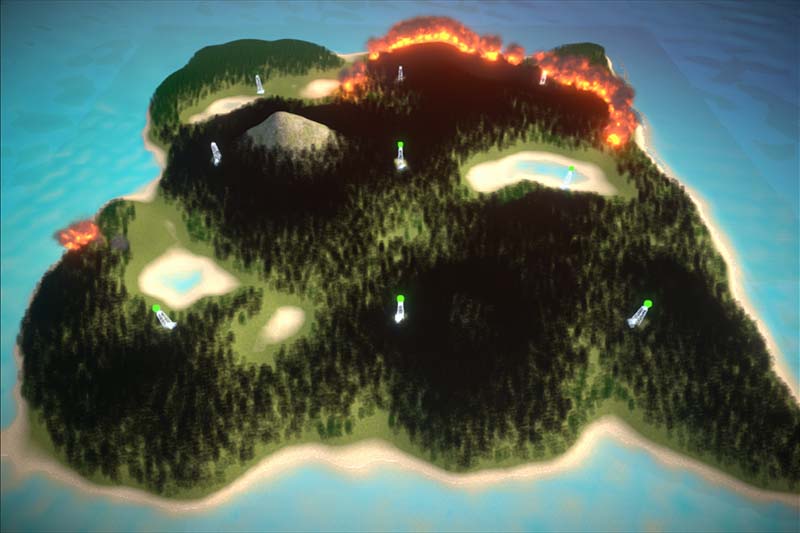}
        \caption{Frame 7}
    \end{subfigure}
    \begin{subfigure}{0.495\textwidth}
        \includegraphics[width=1\linewidth]{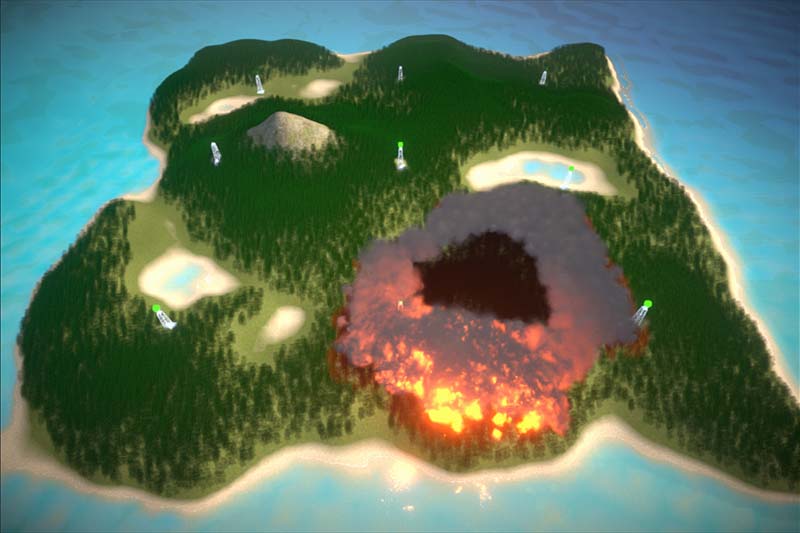}
        \caption{Frame 2}
        \includegraphics[width=1\linewidth]{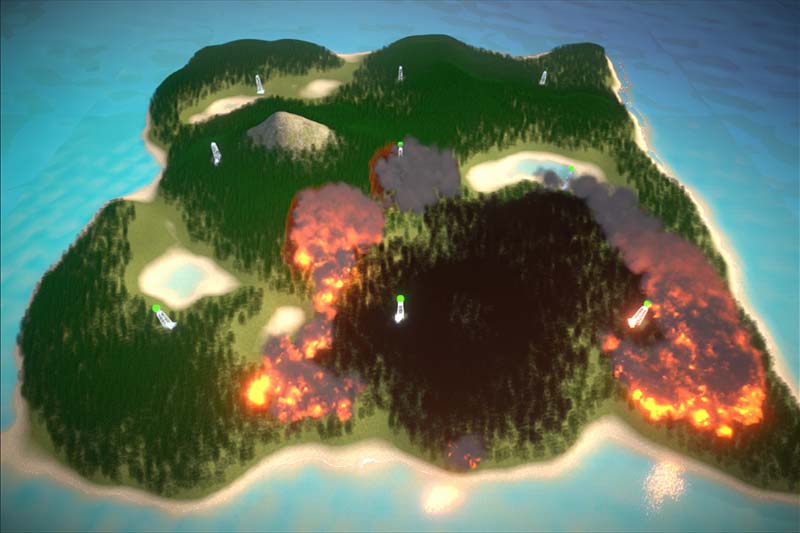}
        \caption{Frame 4}
        \includegraphics[width=1\linewidth]{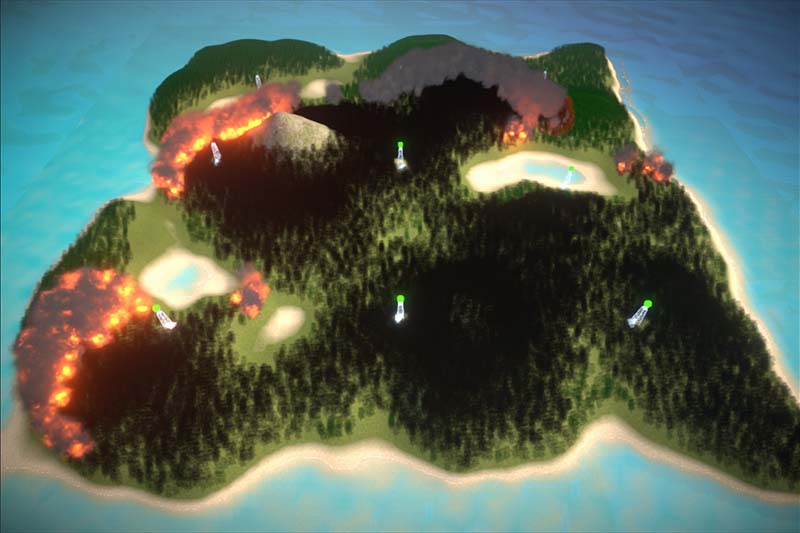}
        \caption{Frame 6}
        \includegraphics[width=1\linewidth]{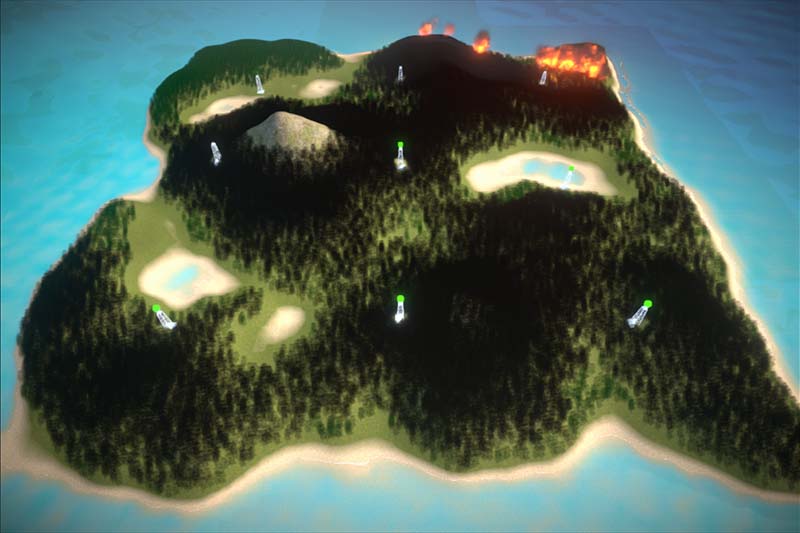}
        \caption{Frame 8}
    \end{subfigure}
\caption{}
\end{figure}

\newpage
\section{Training School Environment Screenshots}
\label{appendix:env-school}
\begin{figure}[H]
\includegraphics[width=1\linewidth]{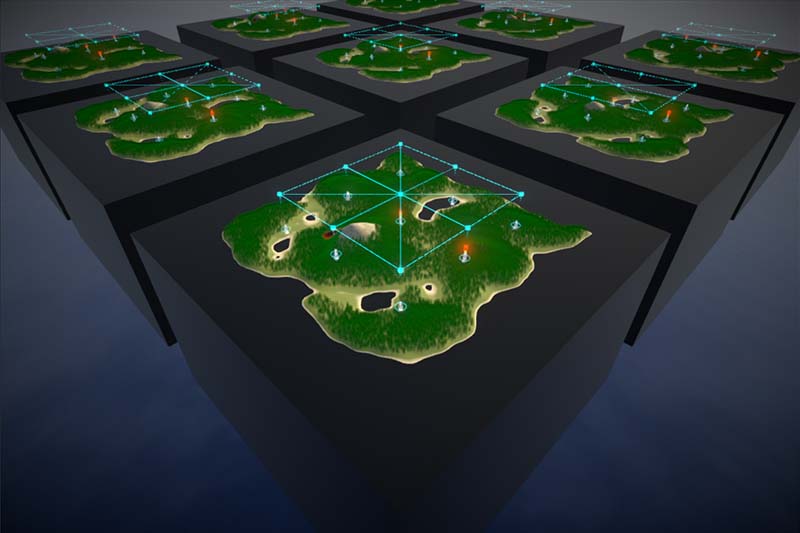}
\caption{Multi-Agent Training School}
\end{figure}
\begin{figure}[H]
\includegraphics[width=1\linewidth]{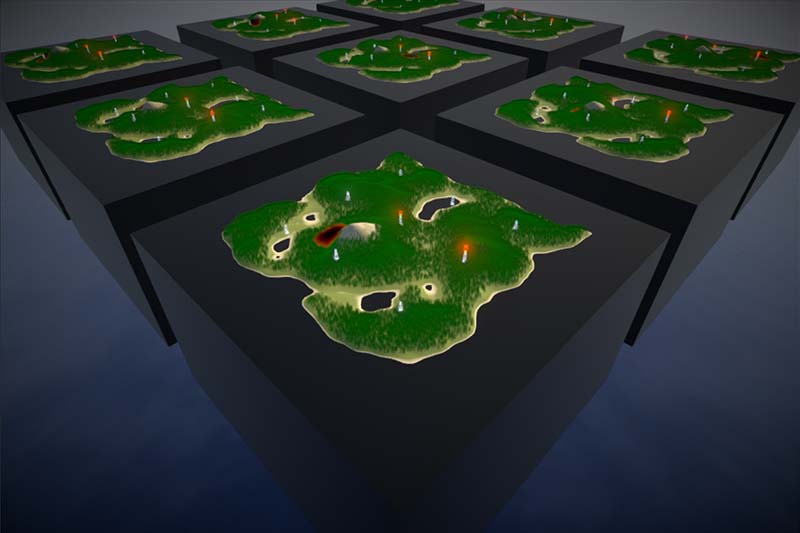}
\caption{Single-Agent Training School}
\end{figure}
\end{document}